\documentclass{article}

\PassOptionsToPackage{numbers, compress}{natbib}

\usepackage[final]{neurips_2024}

\usepackage{microtype}
\usepackage{graphicx}
\usepackage{hyperref}
\usepackage{booktabs} 
\usepackage{adjustbox}  
\usepackage{wrapfig}

\usepackage{amsmath, amssymb, amsthm,bm}
\usepackage{mathtools}
\usepackage{cancel}
\usepackage{empheq}
\usepackage{subcaption}
\usepackage{caption}
\usepackage{nccmath}
\usepackage{algorithm}
\usepackage{algpseudocode}
\usepackage[utf8]{inputenc} 
\usepackage[T1]{fontenc}    
\usepackage{amsfonts}       
\usepackage{nicefrac}       
\usepackage{xcolor}         
\usepackage{multirow}
\usepackage{tabularx}
\usepackage{enumitem}

\usepackage[capitalize,noabbrev]{cleveref}

\usepackage[textsize=tiny]{todonotes}

\DeclarePairedDelimiterX{\infdivx}[2]{[}{]}{%
  #1\;\delimsize\|\;#2%
}

\newcommand{\logsnr}{\alpha}

\newcommand{\vx}{\bm x}

\newcommand{\vz}{\bm x_\logsnr}

\newcommand{\half}{\nicefrac{1}{2}}
\newcommand{\eps}{\bm \epsilon}

\newcommand{\be}{\begin{eqnarray} \begin{aligned}}
\newcommand{\ee}{\end{aligned} \end{eqnarray} }
\newcommand{\benn}{\begin{eqnarray*} \begin{aligned}}
\newcommand{\eenn}{\end{aligned} \end{eqnarray*} }

\newcommand{\epstil}{\tilde{\eps}}
\newcommand{\epshat}{\hat \eps}
\newcommand{\capeps}{\mathcal{E}}

\newcommand{\cx}{\sqrt{\sigma(\logsnr)}}
\newcommand{\cebar}{\sqrt{\sigma(-\bar \logsnr)}}
\newcommand{\cxbar}{\sqrt{\sigma(\bar \logsnr)}}
\newcommand{\ce}{\sqrt{\sigma(-\logsnr)}}

\newcommand{\norm}[1]{{\| #1 \|_2^2 }}

\usepackage{color}

\theoremstyle{plain}

\theoremstyle{definition}

\theoremstyle{remark}

\title{Your Diffusion Model is Secretly a Noise Classifier and Benefits from Contrastive Training}

\author{Yunshu Wu\textsuperscript{1} ,
Yingtao Luo\textsuperscript{2},
Xianghao Kong\textsuperscript{1}, 
Evangelos E. Papalexakis\textsuperscript{1}, 
Greg Ver Steeg\textsuperscript{1} \\
\textsuperscript{1}University of California Riverside, 
\textsuperscript{2}Carnegie Mellon University \\
\texttt{\{ywu380,xkong016,epapalex,gregoryv\}@ucr.edu}, \texttt{yingtaol@andrew.cmu.edu} 
}

\begin{document}

\maketitle

\begin{abstract}
Diffusion models learn to denoise data and the trained denoiser is then used to generate new samples from the data distribution. 
In this paper, we revisit the diffusion sampling process and identify a fundamental cause of sample quality degradation: the denoiser is poorly estimated in regions that are far Outside Of the training Distribution (OOD), and the sampling process inevitably evaluates in these OOD regions.
This can become problematic for all sampling methods, especially when we move to \emph{parallel sampling} which requires us to initialize and update the entire sample trajectory of dynamics in parallel, leading to many OOD evaluations. 
To address this problem, we introduce a new self-supervised training objective that differentiates the levels of noise added to a sample, leading to improved OOD denoising performance. The approach is based on our observation that diffusion models implicitly define a log-likelihood ratio that distinguishes distributions with different amounts of noise, and this expression depends on denoiser performance outside the standard training distribution.
We show by diverse experiments that the proposed contrastive diffusion training is effective for both sequential and parallel settings, and it improves the performance and speed of parallel samplers significantly. \footnote{Code can be found at \url{https://github.com/yunshuwu/ContrastiveDiffusionLoss.git}}
\end{abstract}

\section{Introduction}

Denoising diffusion models~\cite{jaschaneq} achieve state-of-the-art performance on various unsupervised learning tasks and have intriguing theoretical connections to methods like denoising autoencoders~\cite{vincent2011connection}, VAEs~\cite{ho2020denoising}, stochastic differential equations~\cite{mcallester2023mathematics,diffusion_sde}, information theory~\cite{kong2023informationtheoretic}, and score matching~\cite{song2019generative,song2020improved}.
Diffusion models are presented with data samples corrupted by a \emph{forward} dynamical process that progressively adds more Gaussian noise and trained to \emph{reverse} this dynamics or denoise the corrupted samples. Samples are then generated by applying the reverse dynamics on images of pure Gaussian noise to produce high-quality samples from the target distribution.

\begin{figure*}[ht!]
    \centering
    \includegraphics[width=\textwidth]{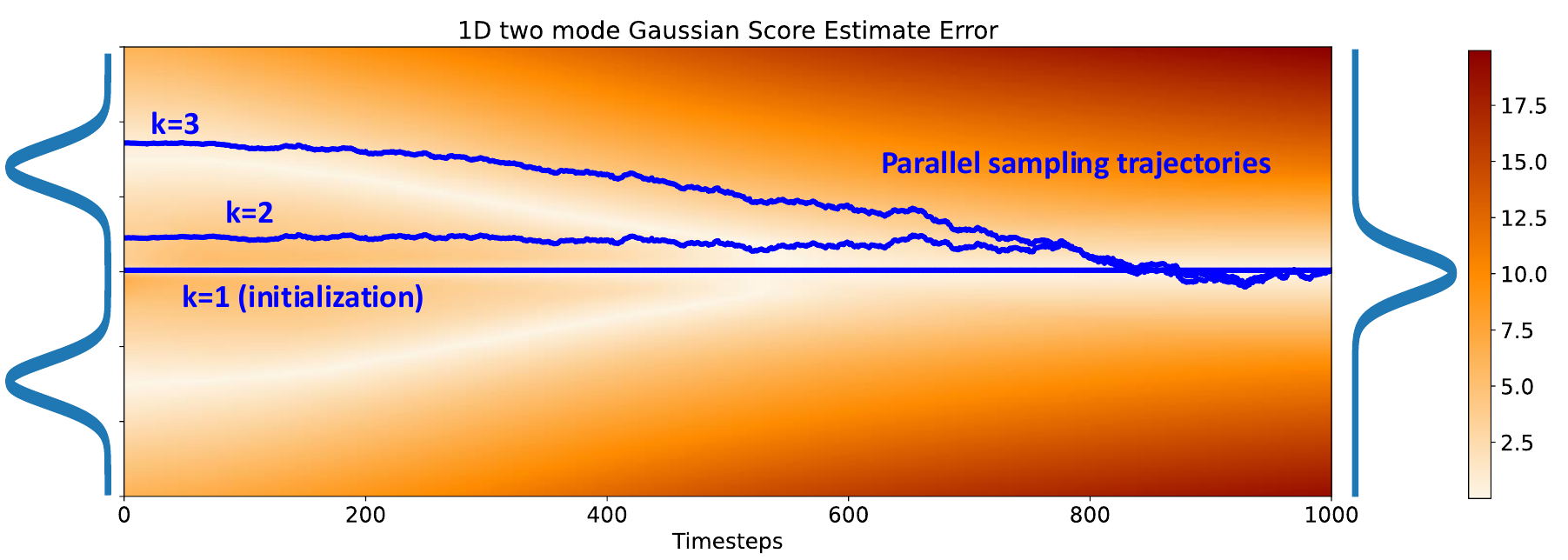}
    \caption{We plot the error in the score estimate for an 1D two mode Gaussian example where diffusion dynamics bridge between a Gaussian and a mixture (see Appendix \ref{sec:1d_gauss_exp}). Regions near the standard forward training data paths have lower error magnitude (light), whereas other areas have higher error magnitude (dark). 
    While sequential samplers adhere as closely as possible to low-error regions, parallel samplers initialize and update the entire sample trajectory (blue trajectories), leading to evaluations in high-error regions. When the sampling trajectory is initialized, most are inevitably in the OOD regions and will update to the low-error regions gradually.}
    \label{fig:main_plot}
    \vspace{-10pt} 
\end{figure*}

The key to the success of diffusion models is the dynamics that gradually bridges the source and a target distribution, but it suffers from slow sampling, as sequentially simulating these dynamics can take thousands of denoising steps for one sample.  
Most recent works attempt to expedite the sequential dynamics by taking fewer, larger steps
~\cite{ddim,karras2022elucidating,lu2022dpm,nichol2021improved}.
However, the complexity of these samplers and the need for expensive sampling hyper-parameter grid searches tailored to specific datasets makes them difficult to generalize. 

\citeauthor{shih2024parallel}~\citeyear{shih2024parallel} suggests a different approach by randomly initializing the entire path of the reverse dynamics and then updating all the steps in the path in parallel. The parallel sampling approach promises to drastically reduce wall-clock time at the cost of increased parallel computation. However, it encounters a problem that has largely gone unnoticed in the sequential sampling literature: while sequential paths sampled during generation are designed carefully to stay as close as possible to the forward paths that add noise to data, the parallel sampler often evaluates in regions far from where the score estimate (the denoiser) is trained, as illustrated in Fig.~\ref{fig:main_plot}. The shading shows that the error of the score estimate is large in these regions, leading to poor performance for parallel samplers. Discretization errors can lead to similar issues, even for standard sequential samplers. 

We propose to improve the training of diffusion models so that the error of the denoiser is reduced in OOD regions, and we hypothesize that this should significantly improve the performance for parallel samplers as they require more OOD evaluations. 
Our approach starts with an unexpected connection: the optimal MSE denoiser that defines the diffusion dynamics \textit{also defines an optimal noise classifier} that distinguishes between samples with different amounts of noise. This provides a useful additional signal for training, because optimizing for the noise classification task involves evaluating the denoiser for one noise level on samples from distributions at different noise levels, while standard MSE optimization only evaluates the denoiser on samples from the matching noisy distribution. Accurate denoiser evaluation in regions that are OOD for standard diffusion training is important for robust sampling dynamics. 

\textbf{Contributions:}
\begin{itemize}[nosep]
    \item We use the information-theoretic formulation of diffusion to draw connections between diffusion, log-likelihood ratio estimation, and classification. This reveals that optimal diffusion denoisers are also implicitly optimal classifiers for predicting the amount of noise added to an image.  
    \item We leverage the noise classifier (via density ratio estimation~\cite{nce}) interpretation to introduce a novel self-supervised loss function for regularizing diffusion model training, which we call the Contrastive Diffusion Loss (CDL). CDL provides training signal in regions that are OOD for the standard MSE diffusion loss. 
    \item We show that CDL improves the trade-off between generation speed and sample quality, and that this advantage is consistent across different models, hyper-parameters, and sampling schemes. The improvement is especially substantial for parallel diffusion samplers~\cite{shih2024parallel} which rely heavily on OOD denoiser evaluations. 
\end{itemize}

\section{Diffusion Model Background: Optimal Denoisers are Density Estimators}\label{sec:dm}

The defining feature of diffusion models is a sequence of distributions that progressively add noise to the data, from which we then learn to recover the original data. 
The (``variance preserving''~\citep{diffusion_sde}) channel that mixes the signal $\vx$ with Gaussian noise is defined as $\vz \equiv \cx \vx + \ce \eps$ with $\eps \sim \mathcal N(0, \mathbb I), \vx \sim p(\vx)$, where $\logsnr$ represents the log of the Signal-to-Noise Ratio (SNR), $p(\vx)$ is the unknown data distribution for $\vx \in \mathbb R^d$, and $\sigma(\cdot)$ is the sigmoid function. 
We define the sequence of intermediate distributions drawn according to this channel with a subscript as $p_\alpha(\vx)$. By definition, we express $\lim_{\logsnr \rightarrow \infty} p_\alpha(\vx) = p(\vx)$ in this paper. 
Note that we use a different scaling convention for noise from ~\cite{karras2022elucidating} and ~\cite{ho2020denoising}, where the former one takes $\vx + \sigma \eps$ as the forward noising channel and the latter one takes $\sqrt{\alpha_t}\vx + \sqrt{1-\alpha_t}\eps$ as the forward noising channel. For further detailed relationships among these scaling conventions, please check App.~\ref{app:sacling_relations}.

The minimum mean square error (MMSE) estimator $\epshat$ for recovering $\eps$ from the noisy channel that mixes $\vx$ and $\eps$ 
can be derived via variational calculus and written as follows. 
\begin{equation}\label{eq:opt}
\epshat(\vz, \logsnr) \equiv \mathbb E_{\eps \sim p(\eps | \vz) }[ \eps] = \arg \min_{\epstil(\cdot, \cdot)} \mathbb E_{p(\eps) p(\vx)} [ \norm{\eps - \epstil(\vx_\logsnr, \logsnr)}]  .
\end{equation}
Sampling from the true posterior is typically intractable, but by using a neural network to approximate the solution to the regression optimization problem, we can get an approximation for $\epshat$. 
From \cite{kong2023informationtheoretic}, we see that log-likelihood can be written \emph{exactly} in terms of an expression that depends only on the MMSE solution to the Gaussian denoising problem, i.e.\
\begin{equation}\label{eq:density_simple} 
    -\log p(\vx) = c + \half \int_{-\infty}^{\infty}  \mathbb E_{p(\eps)} [ \norm{\eps - \epshat(\vx_\logsnr, \logsnr)}]~ d\logsnr. 
\end{equation}
The constant, $c = d/2 \log(2 \pi e) - \frac{d}{2} \int_{0}^{\infty} d\bar\logsnr ~\sigma(\bar\logsnr)$ does not depend on data and will play no role in our approach, as it cancels out in our derivations in Sec. \ref{sec:llr}. 

\section{What Your Diffusion Model is Hiding: Noise Classifiers}\label{sec:llr}

We now introduce our first main result, which shows that diffusion models implicitly define optimal noise classifiers. 
Eq.~(\ref{eq:density_simple}) expresses the probability density of the data directly in terms of the denoising function. 
If we apply Eq.~(\ref{eq:density_simple}) to the noisy distributions that bridge the data and a Gaussian, $p_\zeta(\vx)$, we can see that all mixture densities
can be written in terms of the same optimal denoising function, $\epshat(\cdot, \cdot)$. The complete derivation is presented in App.~\ref{app:mix_density}.
\begin{align}
    -\log p_\zeta(\vx) &= c +  \half \int_{-\infty}^{\infty} d\logsnr ~\mathbb E_{p(\eps)} [ \norm{\eps - b \cdot \epshat(\vx_{\logsnr}, \beta)}]  \label{eq:mix_density} \\
    \vx_{\logsnr} &\equiv \cx \vx + \ce \eps \\
    \beta \equiv \sigma^{-1} &( \sigma(\zeta) \sigma(\logsnr) ),\ 
    b \equiv \sqrt{\sigma(-\logsnr) / \sigma(-\beta)} \label{eq:b}
\end{align}
Intuitively, if we find the optimal denoising function for the data distribution, it may be hypothesized that it can denoise an already \emph{noisy} version of the data distribution. 
Using Eq.~\ref{eq:density_simple}, this directly translates into an expression for density of mixture distributions.
Differences in log likelihoods lead to cancellation of constants, and these Log Likelihood Ratios (LLR) are related to the optimal classifiers~\citep{nce} as we show below. 

To connect LLRs with classification, consider the following generative model. We generate a random binary label $q(y=\pm 1) = 1/2$. Then, conditioned on $y$, we sample from some distribution $q(\vx|y)$. 
Given samples $(\vx, y) \sim q(\vx, y) = q(\vx|y) q(y)$, the Bayes optimal classifier is:
\begin{align}
    q(y|\vx) &= \frac{q(\vx|y) q(y)}{q(\vx)} = \frac{q(\vx|y) q(y)}{q(\vx | y=1) q(y=1) + q(\vx | y=-1) q(y=-1)}  \nonumber \\
    & = 1/(1 + \frac{q(\vx| -y)}{q(\vx|y)} ) = 1/(1+\exp (y (\log q(\vx|y=-1) - \log q(\vx|y=1)))) \nonumber \\ 
     \log q(y|\vx) &= - \log (1+\exp (y \log \frac{q(\vx|y=-1)}{q(\vx|y=1)}))  = -\operatorname{softplus} (y \log \frac{q(\vx|y=-1)}{q(\vx|y=1)}))
     \label{eq:bayes_classifier}
\end{align}
In the second line, because $\forall y, q(y)=1/2$, these constants cancel out. Then we can just expand definitions and re-arrange to write in terms of log probabilities. 

\textbf{Contrastive Diffusion Loss (CDL)~~~~}
Our next contribution is to use the new connection between diffusion denoisers and noise classifiers to define a new training objective. 
We set the distributions $q(\vx|y=1)$ and $q(\vx|y=-1)$ to be two distributions at different noise levels that we can write in terms of the optimal diffusion denoiser from Eq.~\ref{eq:mix_density}. So we have $q(\vx|y=1) \equiv p(\vx)$, the data distribution, and $q(\vx|y=-1) \equiv p_{\zeta}(\vx)$, for some noise level, $\zeta$. 
Then given a sample $(\vx, y) \sim q(\vx, y)$ the per-sample cross-entropy loss for the noise  classifier, Eq.~(\ref{eq:bayes_classifier}), is as follows. 
\begin{align}\label{eq:pointwise_cdl}
\mathcal L_{CDL} 
&= \mathbb E_{q(\vx, y)}\left[ \operatorname{softplus}(y (\log p_\zeta(\vx) - \log p(\vx))) \right]
\end{align}
We can estimate both densities directly from our denoising model using Eq.~(\ref{eq:mix_density}), with the constants canceling out in the process. 
This loss differs significantly from the standard diffusion loss. Intuitively, to distinguish between a sample from the data distribution, $p(\vx)$, versus a noisy version of the data distribution, $p_\zeta(\vx)$, we need to evaluate denoisers on points from both distributions. In standard diffusion training, denoisers at noise level $\zeta$ are only trained on samples from $p_\zeta(x)$. \\
\emph{Limitations:  } We highlight that CDL is more expensive to compute than the standard diffusion loss, significantly increasing the total cost of diffusion model training. Implementation details appear in App.~\ref{app:cdl_implementation} and training cost details appear in App.~\ref{app:cost}.

\textbf{Choice of noise to contrast~~~~} %
Next, let's break the Log-Likelihood Ratio (LLR) term in Eq.~(\ref{eq:pointwise_cdl}) down to see how to choose $\zeta$ to maximize the benefit of CDL. 
Combining Eq.~(\ref{eq:density_simple}) and Eq.~(\ref{eq:mix_density}) we have Eq.~(\ref{eq:llr}), where the constant cancels out.
\begin{align}
    \textit{LLR} = \log p_\zeta(\vx) - \log p(\vx) = \int_{-\infty}^{\infty} d\logsnr ~ \mathbb E_{p(\eps)}[\norm{\eps - \textcolor{blue}{\hat{\eps}(\bm z, \logsnr)}}] - \mathbb E_{p(\eps)}[\norm{\eps - b \textcolor{blue}{\hat{\eps}(\bm z, \beta)}}]  \label{eq:llr} \\
    \text{with:  } \textcolor{blue}{\bm z} \equiv \cx\vx+\ce\eps \nonumber
\end{align}
Note that the input $\vx$ to the LLR term may come from two different distributions, which breaks the standard synchronous denoising pair $(\vx_\logsnr, \logsnr)$ into asynchronous. When it's from data distribution $\vx \sim p(\vx)$, \textcolor{blue}{$\bm z = \bm z_\logsnr$}; and when it's from some noisy data distribution $\vx \sim p_\zeta(\vx)$, \textcolor{blue}{$\bm z = \bm z_\beta$}. 

From Eq.~(\ref{eq:llr}) we see that $\hat{\eps}(\cdot,\cdot)$ is trained on four pairs: $(\bm z_\logsnr, \logsnr)$, $(\bm z_\beta, \beta)$, $(\bm z_\beta, \logsnr)$ and $(\bm z_\logsnr, \beta)$, where $\beta \equiv \sigma^{-1}(\sigma(\logsnr)\sigma(\zeta)) < \min(\logsnr,\zeta)$ (Eq.~\ref{eq:b}). During standard training, only the first two pairs are trained (Eq.~\ref{eq:opt}). 
This means that our CDL objective trains the denoiser to perform correctly even for samples from distributions that are noisier or cleaner than the specified noise level (a pair like $(\bm z_\beta, \logsnr)$ or $(\bm z_\logsnr, \beta)$). 
This can be useful for both sequential and parallel sampling settings. 
During sequential sampling, extra error noise added due to discretization errors can be corrected by the denoiser trained with CDL. 
As for parallel sampling, CDL helps with evaluations on asymmetric pairs $(\bm z_\beta, \logsnr)$ or $(\bm z_\logsnr, \beta)$ which we refer to OOD regions for standard diffusion loss. 

In practice, diffusion training pipelines are highly tuned on popular datasets like CIFAR10 and ImageNet, so the amplitude of discretization errors during sampling is small, meaning that errors won't nudge points too far away from the true trajectory. 
Therefore, when evaluating CDL objective, we sample some large-valued $\zeta$s, which corresponds to classifying only small differences in noise levels. 
Empirically we find that $\zeta \sim \textit{Uniform}[6,15]$ or $\zeta \sim \textit{logistic}[6,15]$ performed equally good. 

\textbf{Denoising, sampling dynamics, and the score connection~~~~}\label{sec:score-denoise}
We have focused so far on denoising and density estimation, but we now want to connect this discussion to the primary use case for diffusion models and the focus of Sec.~\ref{sec:sampling}, \emph{sampling}. There are many choices in how to implement sampling dynamics~\cite{karras2022elucidating}, but all of them rely on the \emph{score function}, $\nabla_x \log p_{\logsnr}(\vx)$. The score function points toward regions of space with high likelihood, and by slowly transitioning (or annealing), from the score function of a noisy distribution to one closer to the data distribution, we can build reliable sampling dynamics. To connect denoisers with sampling we must show that a denoising function, $\epshat$, that is optimal according to Eq.~\ref{eq:opt} also specifies the score function. 
\begin{align}\label{eq:score}
\nabla_x \log p_{\logsnr}(\vx) = - \frac{\epshat(\vx, \logsnr)}{\ce}
\end{align}
The derivation is straightforward and is given in Appendix  \ref{app:score}.

\section{Sequential and Parallel Sampling with Diffusion Models} \label{sec:sampling}

Sampling dynamics are typically presented in terms of a stochastic process $\{\vx_t\}_{t=1}^T$ with timestep, $t$, rather than in terms of log SNR, $\logsnr$. We will denote $\vx_t \equiv \vx_{\logsnr(t)}, p_t(\vx) \equiv p_{\logsnr(t)}(\vx)$, to connect to our previous notation, with $\logsnr(t)$ representing a monotonic relationship described in App.~\ref{app:sacling_relations}. 
Note that decreasing $\log$-SNR $\logsnr$ corresponds to increase timestep $t$, since smaller $\log$-SNR means there is more noise added to the data. 

The general form of sampling dynamics is a process of slowly transitioning from samples of a simple and tractable distribution to the target distribution. 
Specifically, start with an isotropic Gaussian $\vx_T \sim \mathcal N(0,\mathbb I)$, the sampler steps through a series of intermediate distributions with noise levels $\{T, T-1, \dots, 1 \}$ following the score estimates. 
Many works ~\cite{diffusion_sde,karras2022elucidating} interpret diffusion models as stochastic differential equations (SDEs). 
The forward process is in the form of
\begin{equation}
    d\vx_t = \underbrace{f(\vx_t,t)}_{\text{drift}~s} dt + \underbrace{g(t)}_{\text{diffusion}} d \bm w_t, ~~~~~~~~\vx_0 \sim p(\vx) \label{eq:forward_sde}
\end{equation}
where $\bm w_t$ is the standard Wiener process/Brownian motion, $f$ and $g$ are drift coefficient and diffusion coefficient of $\vx_t$ separately. 
The reverse process of Eq.~\ref{eq:forward_sde} is then used to generate samples
\begin{equation}
    d\vx_t = \underbrace{( f(\vx_t,t) \vx - g^2(t) \nabla_{\vx} \log p_t(\vx) )}_{\text{drift}~s}dt + \underbrace{g(t)}_{\text{diffusion}} d \bm w_t, ~~~~~~~~\vx_T \sim p(\vx) \label{eq:reverse_sde}
\end{equation}

Depending on choices for $f,g$, we can get either a stochastic or ordinary (deterministic) differential equation. 
Either way, numerical differential equation solvers are used to approximate the true dynamics. 
The solver introduces discretization errors at each step, causing the trajectory to deviate into the OOD region where the score (or denoiser) is poorly estimated, further compounding the errors. 
More discretization steps reduce accumulated error and leads to better sample quality, at the expense of more sequential computation. As a result, a significant limitation of diffusion models is that they require many iterations to produce high quality samples.

\textbf{Sequential Sampling~~~~}
The influential diffusion sampler DDPM ~\cite{ho2020denoising} iterates over thousands of discretization steps in simulating the dynamics. 
Recently, many sequential sampling methods have been developed to take fewer and larger steps while introducing less error ~\cite{ddim,nichol2021improved,karras2022elucidating}.
Specifically, ~\citeauthor{karras2022elucidating}~\citeyear{karras2022elucidating} studies the curvature shape of SDE/ODE trajectory and suggests a discretization technique where the resulting tangent of the solution trajectory always points towards the denoiser output.
However, speeding up the sequential sampling sacrifices generation quality.
Furthermore, the SOTA sequential samplers \cite{lu2022dpm,karras2022elucidating,ddim} require hyperparameter tuning and grid search on specific datasets, which poses challenges to the generalization of these samplers to other datasets. 

\begin{figure}[h]
    \centering
    \includegraphics[width=0.4\linewidth]{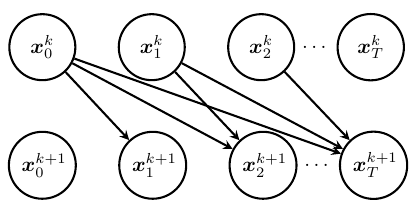}
    \caption{The computation graph of Picard iteration for parallel sampling ~\cite{shih2024parallel}}
    \label{fig:para_diagram}
    \vspace{-10pt}
\end{figure}

\textbf{Parallel Sampling~~~~}
\citeauthor{shih2024parallel}~\citeyear{shih2024parallel} explores a parallel sampling scheme, where the entire reverse process path is randomly initialized and then all steps in the path are updated in parallel. 
Parallel sampling is based on the method of Picard iteration, an old technique for solving ODEs through fixed-point iteration. An ODE is defined by a drift function $s(\vx,t)$ and initial value $\vx_0$. In the integral form, the value at time $t$ can be written as
\begin{equation*}
    \vx_t = \vx_0 + \int_{0}^{t} s(\vx_u,u) \, du
\end{equation*}
In other words, the value at time $t$ must be initial value plus the integral of the derivative along the path of the solution. This formula suggests a natural way of solving the ODE by starting with a guess of the solution $\{ \vx_t^{k+1}:0\leq t \leq 1 \}$ at initial iteration $k=0$, and iteratively refining by updating the value at every time $t$ until convergence \footnote{For detailed convergence proof, we refer to \citeauthor{shih2024parallel}~\citeyear{shih2024parallel} section 3.}
\begin{flalign} \label{eq:picard_int}
    \text{\textbf{(Picard Iteration)}} && \vx_t^{k+1} = \vx_0^k + \int_{0}^{t} s(\vx_u^k, u) \, du &&
\end{flalign}
To perform Picard iterations numerically, which is shown in Fig.~\ref{fig:para_diagram}, we can write the discretized form of Eq.~\ref{eq:picard_int} with step size $1/T$, for $t \in [0, T]$:
\begin{equation} \label{eq:picard_sum}
    \vx_t^{k+1} = \vx_0^k + \frac{1}{T}\sum_{i=0}^{t-1} s(\vx_i^k, i/T)
\end{equation}
We see that the expensive computations $\{ s(\vx_i^k,i/T): i \in [0,T) \}$ can be performed in parallel. 
After some number of Picard iterations, the error difference between two iterates $\norm{\vx_t^{k+1} - \vx_t^k}$ drops below some convergence threshold. 
This converged trajectory, $\vx_t^*$, should be close to the sequential sampler trajectory. 
Looking at the example in Fig.~\ref{fig:main_plot}, we show the trajectories of three iterations $k=1,2,3$. 
The trajectories before convergence are consistently appearing in the regions with high score error.

\section{Experiments}\label{sec:exp}

We sample from models fine-tuned on Contrastive Diffusion Loss (CDL) via both parallel and sequential diffusion samplers across a variety of generation tasks, including complex low-dimensional manifold 2D synthetic data and real-world image generation.
Our results demonstrate that employing CDL as a regularizer in models trained with standard diffusion loss enhances density estimation and sample quality while also accelerating convergence in parallel sampling. 
All sampling tests are done on A6000 GPUs. 
We visualize CDL image generation examples in App.~\ref{app:samplers}.

\textbf{Training configuration~~~~}
Our method is architecture-agnostic. 
In synthetic experiments, we adopt a simple MLP architecture with positional encoding for timesteps~\footnote{Architecture adopted from: \href{https://github.com/Jmkernes/Diffusion}{https://github.com/Jmkernes/Diffusion}}, as it is one of the most versatile models in the literature. 
In real-world experiments, we consider the standard diffusion loss with two training configurations: DDPM by ~\citeauthor{ho2020denoising}~\citeyear{ho2020denoising} and EDM by ~\citeauthor{karras2022elucidating}~\citeyear{karras2022elucidating}. For more details on model training, data split, and hyper-parameters, please refer to App.~\ref{app:model_training}.

\textbf{Generation quality metrics~~~~}
For real-world data, our intrinsic metric is Fre\'chet Inception Distance (FID) score~\cite{heusel2017gans}. 
The number of images we generated for FID computation follows their baseline models' FID settings, and the FID scores are computed between $5,0000$ generated images and all available real images. 

For synthetic data, to measure how well the generated samples resemble samples from the ground truth distribution, we use the (unbiased) kernel estimator of the squared Maximum Mean Discrepancy (MMD), with Gaussian kernel with bandwidth set empirically as described in App.~\ref{app:mmd}.

\textbf{Sampling speed metrics~~~~} 
We adopt the following three metrics: 
(1) Neural function evaluations (NFE) for all settings, i.e. how many times the denoiser is evaluated to produce a sample; 
(2) For the parallel setting, we report the number of parallel Picard Iterations;
(3) Furthermore for the parallel setting, the wall-clock time is reported. 
While parallel sampling can use fewer total iterations and less wall-clock time than a sequential sampler, this may come at the cost of an increase in the total number of function evaluations. This gap is called the algorithm inefficiency.
In the subsequent section, we use contrastive diffusion loss as a training regularizer for standard diffusion losses and refer to the corresponding models as CDL-regularized models.

\subsection{Parallel Sampling}

In the parallel setting, we use Parallel DDPM sampler~\cite{shih2024parallel} with 1000-step diffusion sampling. 
And for synthetic experiment, to reflect sampling speed by only number of Picard iteration and wall-clock time, we don't use the sliding window technique, and the 2D data is small enough to fit the whole sampler trajectory in GPU memory. While for real-world experiment, sliding window is still applied. 

\textbf{Synthetic Dataset~~~~}
We consider the 2D Dino dataset~\cite{dino_data}, characterized by its highly nonlinear density concentrated on a low-dimensional manifold. 
For baselines, we employ the standard DDPM loss ~\cite{ho2020denoising}, as all standard diffusion losses similarly minimize a sum of MSE losses between the actual and estimated denoisers.
Both CDL-regularized and DDPM-objective-trained models are trained with a MLP where timestep is encoded by positional encoding. We train it for 2000 epochs to ensure convergence and check the training and validation loss curve to avoid overfitting.

\begin{figure*}[h]
    \centering
    \begin{minipage}{0.56\textwidth}
        \centering
        \captionsetup{font=scriptsize}

        \begin{subfigure}{0.32\textwidth}
            \includegraphics[width=\linewidth]{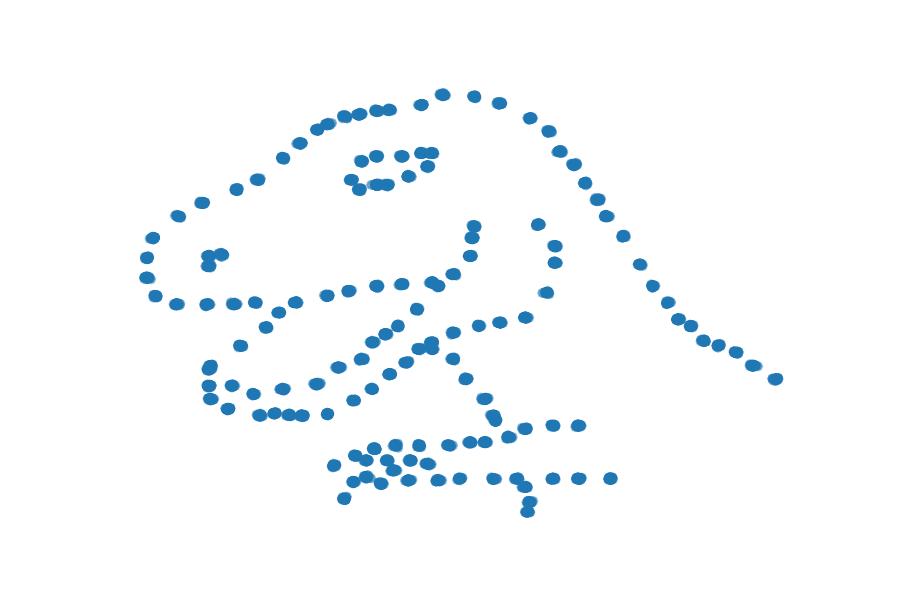}
            \caption{Ground-truth dino.}
            \label{fig:dino_gt}
        \end{subfigure}
        \begin{subfigure}{0.32\textwidth}
            \includegraphics[width=\linewidth]{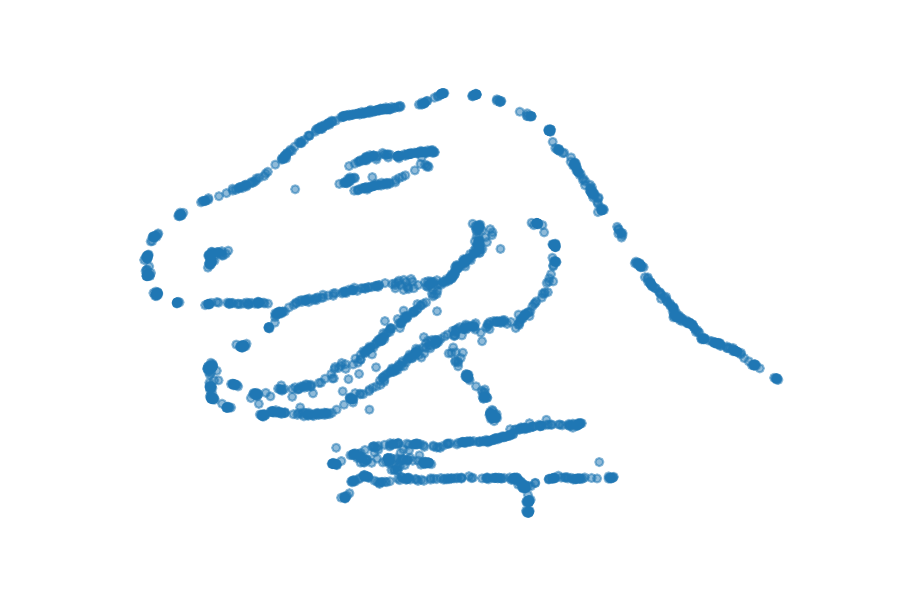}
            \caption{Using CDL loss.}
            \label{fig:dino_cdl}
        \end{subfigure}
        \begin{subfigure}{0.32\textwidth}
            \includegraphics[width=\linewidth]{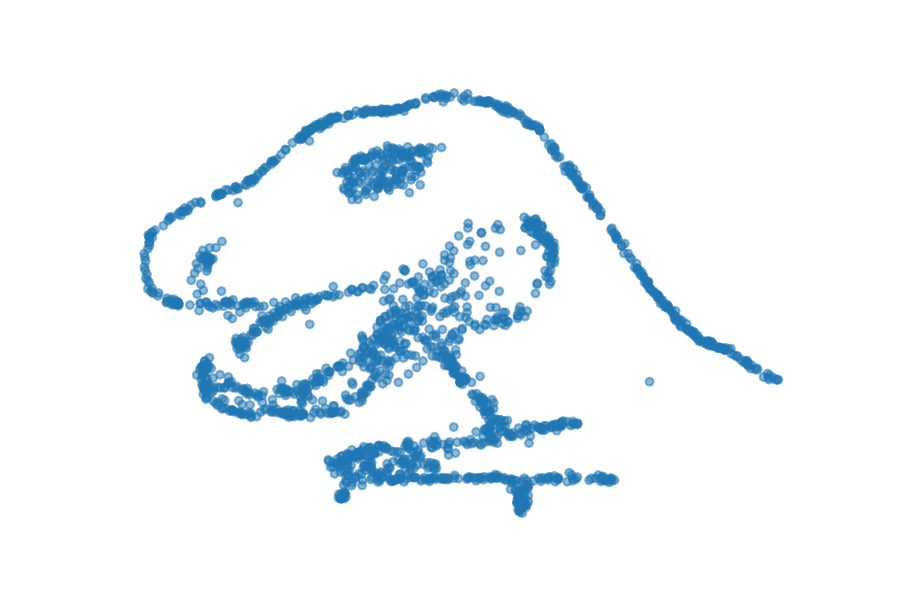}
            \caption{Using DDPM loss.}
            \label{fig:dino_ddpm}
        \end{subfigure}
        \captionof{figure}{Parallel DDPM sampler generated Dino data. Comparing to Dino sampled from DDPM loss, CDL-loss sampled Dino has better sample quality and density estimate around hard areas. }
        \label{fig:dino_samples}

        \scriptsize
        \begin{tabular}{l|cccc}
            \toprule
            Model Loss & MMD & \#Iter &  NFE & Time (ns) \\
            \midrule
            DDPM & 0.0031 &  36 & 14,397 & 1,870 \\
            CDL & \bm{$0.0012$} &  \bm{$27$} & \bm{$13,983$} & \bm{$1,368$} \\
            \bottomrule
        \end{tabular}
        \captionof{table}{Parallel DDPM sampling speed results. We generate $2,000$ samples. Here we set MMD threshold$=0.002$, and \#Iter refers to number of picard iterations till MMD threshold. Both NFE and Time are counted till parallel convergence.}  
        \label{tab:dino_stats}
    \end{minipage}
    \hfill
    \begin{minipage}{0.43\textwidth}
        \centering
        \captionsetup{font=scriptsize}
        \includegraphics[width=\linewidth]{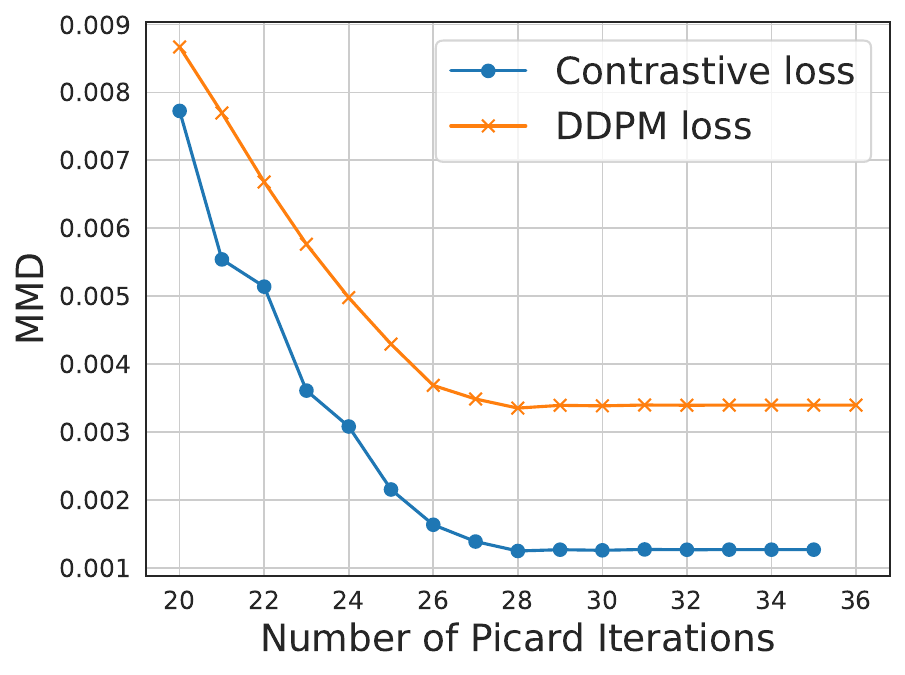}
        \captionof{figure}{Parallel sampling MMD plot. We can see that the CDL-regularized model and DDPM model converge themselves at 35 and 36 Picard iterations, separately.} 
        \label{fig:2D_dino_mmd}
    \end{minipage}
    \vspace{-13pt}
\end{figure*}

In Fig.~\ref{fig:dino_samples}, it is clearly demonstrated that the parallel generated samples from the CDL-regularized model is much better than the model trained only with the standard DDPM loss, especially around the chin, eyes and paws, where the manifolds are close to each other and difficult to distinguish and learn.
From the MMD plot, we see that comparing to the baseline curve trained only with standard DDPM loss, the CDL-regularized curve converges faster with smaller number of Picard iterations and better sample quality (lower MMD scores). 
Table.~\ref{tab:dino_stats} shows the sampling speed results, from where we see that CDL-regularized model converges faster with lower final MMD and better sample quality.

\textbf{Real-world Datasets~~~~}
We select 
“DDPM++ cont. (VP)” and “NCSN++ cont. (VE)” models by~\cite{karras2022elucidating} trained on CIFAR-10 at $32\times32$, unconditional FFHQ, and unconditional AFHQv2 ~\cite{krizhevsky2009learning,karras2019style,choi2020stargan} as baselines, comparing to the corresponding CDL-regularized models. 
We adopt the pre-trained models from \citeauthor{ho2020denoising}~\citeyear{ho2020denoising}\footnote{\url{https://github.com/pesser/pytorch_diffusion}} and ~\citeauthor{karras2022elucidating}~\citeyear{karras2022elucidating}\footnote{\url{https://github.com/NVlabs/edm}}.
More experimental results can be found in App.~\ref{app:model_training}. 
As shown in Tab.~\ref{tab:parallel_sampling}, CDL-regularized models always outperformed baselines with respect to FID scores.

\begin{table}[h]
    \centering
    \begin{adjustbox}{center}
    \small
    \begin{tabular}{@{}l|cccc@{}} 
        \toprule
        \multirow{2}{*}{Models} & \multicolumn{2}{c}{CIFAR-10 at 32x32} & AFHQv2 64x64 & FFHQ 64x64 \\
        \cmidrule(lr){2-5}
         & unconditional & conditional & unconditional & unconditional \\
        \midrule
        VP & $3.24 \pm 0.02$ & $2.93 \pm 0.02$ & $2.95 \pm 0.03$ & $3.67 \pm 0.04$ \\
        CDL-VP & \bm{$2.51 \pm 0.01$} & \bm{$2.41 \pm 0.01$} & \bm{$2.91 \pm 0.02$} & \bm{$3.33 \pm 0.03$} \\
        \midrule
        VE & $3.00 \pm 0.01$ & $2.76 \pm 0.01$ & $2.98 \pm 0.03$ & $3.65 \pm 0.02$ \\
        CDL-VE & \bm{$2.38 \pm 0.01$} & \bm{$2.25 \pm 0.02$} & \bm{$2.93 \pm 0.01$} & \bm{$3.29 \pm 0.02$} \\
        \bottomrule
    \end{tabular}
    \end{adjustbox}
    \caption{Evaluating FID score (lower is better) of parallel DDPM sampler on real-world datasets using $5,0000$ samples. For reported FID scores, we run three sets of random seeds and reported the average with uncertainty.}
    \label{tab:parallel_sampling}
    \vspace{-13pt}
\end{table}

\subsection{Sequential Sampling}

While CDL clearly improves parallel sampling quality and convergence speed, we also show that it improves the trade-off between generation speed and sample quality in the sequential setting. 
As for sequential diffusion sampling choices, we consider the DDPM sampler from~\citeauthor{ho2020denoising}~\citeyear{ho2020denoising}, and both the deterministic and stochastic samplers from~\citeauthor{karras2022elucidating}~\citeyear{karras2022elucidating}. To ensure fair comparisons, we adopt the original sampling hyper-parameter settings for all baselines. 

\textbf{Deterministic samplers~~~~}
For FID test, we follow the exact sampling settings outlined in \citeauthor{karras2022elucidating}~\citeyear{karras2022elucidating} for each dataset.
FID scores are reported in Tab.~\ref{tab:ODE_table}, for sequential deterministic EDM samplers, CDL objective ensures that the generation quality is consistently similar or better.

\begin{figure}[t]
    \centering
    \includegraphics[width=0.99\linewidth]{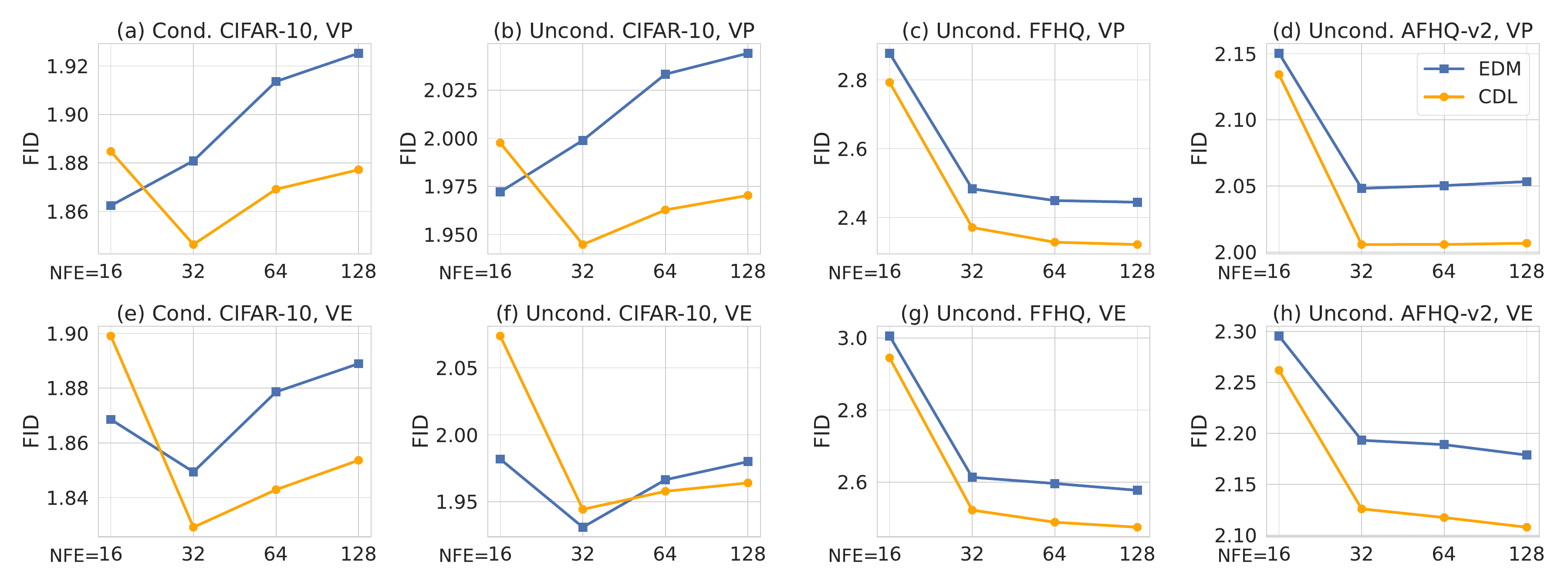}
    \caption{The FID comparison between our CDL and the baseline EDM in the deterministic sampler experiment.}
    \label{fig:ode_only}
    \vspace{-13pt}
\end{figure}

\begin{table}[h]
    \centering
    \begin{adjustbox}{center}
    \small
    \begin{tabular}{@{}l|cccc@{}}  
        \toprule
        \multirow{2}{*}{Models} & \multicolumn{2}{c}{CIFAR-10 at 32x32} & AFHQv2 64x64 & FFHQ 64x64 \\
        \cmidrule(lr){2-5}
         & unconditional & conditional & unconditional & unconditional \\
         \midrule
        VP      & $2.00 \pm 0.02$ & $1.84 \pm 0.02$ & $2.04 \pm 0.00$ & $2.38 \pm 0.01$ \\
        CDL-VP  & \bm{$1.99 \pm 0.04$} & \bm{$1.82 \pm 0.03$} & \bm{$2.00 \pm 0.00$} & \bm{$2.29 \pm 0.02$} \\
        \midrule
        VE      & $2.01 \pm 0.01$ & $1.81 \pm 0.01$ & $2.17 \pm 0.00$ & $2.56 \pm 0.03$ \\
        CDL-VE  & $2.01 \pm 0.01$ & $1.81 \pm 0.01$ & \bm{$2.11 \pm 0.01$} & \bm{$2.47 \pm 0.02$} \\
        \midrule
        NFE (EDM/CDL) & 35 & 35 & 79 & 79 \\ \bottomrule
        \bottomrule
    \end{tabular}
    \end{adjustbox}
    \caption{Evaluating sequential deterministic EDM samplers generation quality. For reported FID scores, we run three sets of random seeds and reported the average with uncertainty.}
    \label{tab:ODE_table}
    \vspace{-13pt}
\end{table}
In principle, increasing NFE has the potential to decrease the overall discretization errors, consequently leading to improved sample quality. However, in practice we observed an unusual behavior\footnote{The same issue is also reported in \url{https://github.com/NVlabs/edm/issues/4}} with the Karras deterministic sampler -- as NFE increases, the FID score deteriorates (Fig.~\ref{fig:ode_only}). 
In contrast to EDM models, CDL-regularized models exhibit a more stable FID score. This partially resolves the deterministic sampler sensitivity while improving the quality.

\textbf{Stochastic samplers~~~~}
In practice, stochastic samplers often yield superior performance compared to deterministic ones.
However this is not true in \citeauthor{karras2022elucidating}~\citeyear{karras2022elucidating}: stochastic samplers outperform deterministic ones only on challenging datasets, for simpler datasets, the introduction of stochasticity not only fails to enhance performance but exhibits image degradation issues, characterized by a loss of detail. 
They attribute this phenomenon to L2-trained denoisers excessively removing noise at each step (always remove more than it should), and propose to slightly increase the standard deviation ($S_\text{noise}$)~\footnote{We refer to \citet{karras2022elucidating} for details} of newly added noise to 1.007. 
We argue that this approach may not totally resolve the issue and instead complicates the hyperparameter grid search process by introducing an additional parameter, $S_\text{noise}$. Also, this $S_\text{noise}$ logically serves the same function as another hyperparameter $\gamma_i$, where both of them control the amount of noise to add to reach a higher noise level. 

In this experiment, we conducted two stochastic sampling configurations for our baseline EDM-trained models. The first configuration, referred to as EDM-opt, operated at the EDM optimal setting with $S_\text{noise}=1.007$. The second, named as EDM-sub-opt, used a setting with $S_\text{noise}=1.00$, effectively disabling $S_\text{noise}$. As for CDL configuration, we exclusively examined the scenario with $S_\text{noise}=1.00$ to determine whether CDL could address the problem of excessive noise removal.

\begin{figure}[h]
    \centering
    \includegraphics[width=0.5\linewidth]{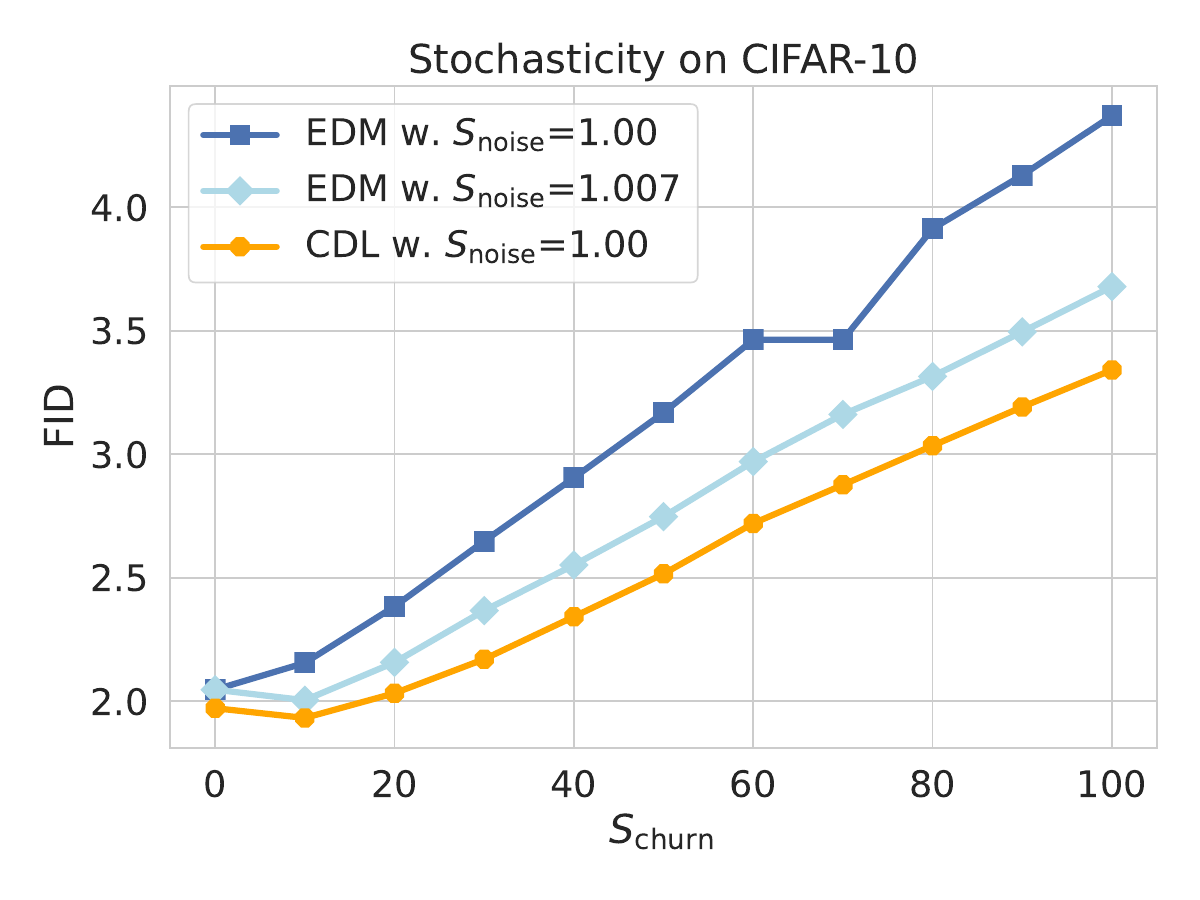}
    \caption{The FID comparison between our CDL and the baselines EDM in the stochastic sampler experiment on CIFAR-10. CDL's performance is strictly better for all $S_{\text{churn}}$, outperforming the optimal setting of EDM which inflates the standard deviation $S_{\text{churn}}$ of the newly added noise.}
    \label{fig: sde_cifar10}
    \vspace{-13pt}
\end{figure}

The results, as visualized in Figure~\ref{fig: sde_cifar10}, indicate that CDL outperforms EDM in both 
$S_\text{noise}$ configurations. Notably, CDL not only improves upon the EDM-sub-opt configuration (dark blue line) but also surpasses the performance of the EDM-opt configuration (light blue line), even at its optimal setting. 
This not only demonstrates that CDL robustly provides a better sample quality, but also suggests that CDL can eliminate the need for the hyperparameter $S_\text{noise}$. This reduction enables a more efficient grid search for the optimal EDM sampling settings, potentially enhancing the practicality of using such a sampler for other applications.

\section{Related Work}

The generative modeling trilemma~\cite{xiao2021tackling} seeks generative models that produce samples (i) quickly, (ii) with high quality, and (iii) good mode coverage. Diffusion models excel at the latter two but a large amount of research has attempted to address the problem of slow sampling speed. 
From the inception of diffusion models~\cite{jaschaneq}, dynamics has been at the forefront, so most work has focused on the interpretation of the dynamics as a differential equation that can be sped up through accelerated numerical solvers~\cite{ddim,diffusion_sde,zhang2021diffusion,jolicoeur2021gotta,liu2022pseudo}. Our approach is compatible with any of these approaches as we are sampler agnostic, seeking only to improve the input to the sampler, which is the denoiser or score function estimator, through regularization during the diffusion model training. 
A separate line of work instead attempts to distill a diffusion model into a faster model that achieves similar sampling quality~\cite{salimans2021progressive,song2023consistency,meng2023distillation,watsonlearning,xiao2021tackling}. In principle, these methods could also benefit from distilling based on a more robust base diffusion model trained with CDL. 

Diffusion models admit a surprisingly diverse array of mathematical perspectives, like variational perspectives~\cite{ho2020denoising,huang2021variational,vdm}, differential equations~\cite{diffusion_sde,mcallester2023mathematics}, and nonequilibrium thermodynamics~\cite{jaschaneq}. Our approach is mostly inspired by connections between the information-theoretic perspective~\cite{kong2023informationtheoretic,konginterpretable} and the score matching perspective~\cite{song2019generative,song2019sliced,song2020improved,hyvarinen2005estimation,vincent2011connection}. In particular, we point out that score function estimates in traditional diffusion training are sub-optimal, and the information-theoretic perspective leads to a new objective (CDL) that can improve the score estimate. 

%

While previous diffusion models focus on log-likelihood estimation, we consider a different approach based on density ratio estimation and noise contrastive estimation \cite{nce,cpc}, which inspired several notable developments in machine learning~\cite{simclr,clip}. A few works have considered contrastive learning inspired modifications to diffusion either to enforce multimodal data relationships~\cite{lee2023codi,zhu2022discrete}, for style transfer~\cite{yang2023zero}, or for guidance during generation~\cite{ouyang2022improving}, but none use the diffusion model as a noise classifier to improve diffusion training as we do.  
Most similar to our approach are methods that use Density Ratio Estimation (DRE) to estimate a ratio between the data density and some simple noise distribution. The density ratio can be estimated by learning to contrast between data samples and samples from the noisy distribution ~\cite{nce,Sugiyama_Suzuki_Kanamori_2012}. Recent work generalized the idea to consider classifying between samples along a sequence of distributions between source and target~\cite{rhodes2020telescoping,choi2021density}. 
Our contribution is to relate this approach to diffusion models by noting that standard diffusion models implicitly already implement the required classifiers for distinguishing distributions on the path from the data distribution to a Gaussian distribution. Concurrent work makes a similar connection but while we focus on improving diffusion models by interpreting them as noise classifiers, \cite{yadin2024classification} focused on the converse perspective, improving density ratio estimation by interpreting DREs as denoisers.

\section{Conclusion}

In this paper, we introduced a novel connection between diffusion models and optimal noise classifiers. 
While this relationship has a variety of potential applications that could be explored in future work, we used the connection to propose a new self-supervised loss regularizer for diffusion models, the Contrastive Diffusion Loss (CDL). 
CDL reduces the error of the learned denoiser in regions that are OOD for the standard loss.
We showed that CDL improves the robustness of diffusion models across all types of sampling dynamics, and leads to significant speed-ups for a promising new generation of parallel samplers. 

\section*{Acknowledgements}
This research was partially supported by the National Science Foundation under grant no. IIS 1901379.

\textbf{Broader Impacts:  }
This paper aims to innovate on the methodology of diffusion models. We anticipate no direct potential societal consequences of our work, as the main focus of this work is in theory and algorithm design. 
However, it is important to acknowledge the potential risks associated with diffusion models, as misuse can contribute to the spread of disinformation and deepfakes.

{
\small
\bibliographystyle{plainnat}
\bibliography{diffusion}

\begin{thebibliography}{43}
\providecommand{\natexlab}[1]{#1}
\providecommand{\url}[1]{\texttt{#1}}
\expandafter\ifx\csname urlstyle\endcsname\relax
  \providecommand{\doi}[1]{doi: #1}\else
  \providecommand{\doi}{doi: \begingroup \urlstyle{rm}\Url}\fi

\bibitem[Chen et~al.(2020)Chen, Kornblith, Norouzi, and Hinton]{simclr}
Ting Chen, Simon Kornblith, Mohammad Norouzi, and Geoffrey Hinton.
\newblock A simple framework for contrastive learning of visual representations.
\newblock In \emph{International conference on machine learning}, pages 1597--1607. PMLR, 2020.

\bibitem[Choi et~al.(2021)Choi, Meng, Song, and Ermon]{choi2021density}
Kristy Choi, Chenlin Meng, Yang Song, and Stefano Ermon.
\newblock Density ratio estimation via infinitesimal classification.
\newblock \emph{arXiv preprint arXiv:2111.11010}, 2021.

\bibitem[Choi et~al.(2020)Choi, Uh, Yoo, and Ha]{choi2020stargan}
Yunjey Choi, Youngjung Uh, Jaejun Yoo, and Jung-Woo Ha.
\newblock Stargan v2: Diverse image synthesis for multiple domains.
\newblock In \emph{Proceedings of the IEEE/CVF conference on computer vision and pattern recognition}, pages 8188--8197, 2020.

\bibitem[Gutmann and Hyv{\"a}rinen(2010)]{nce}
Michael Gutmann and Aapo Hyv{\"a}rinen.
\newblock Noise-contrastive estimation: A new estimation principle for unnormalized statistical models.
\newblock In \emph{Proceedings of the thirteenth international conference on artificial intelligence and statistics}, pages 297--304. JMLR Workshop and Conference Proceedings, 2010.

\bibitem[Heusel et~al.(2017)Heusel, Ramsauer, Unterthiner, Nessler, and Hochreiter]{heusel2017gans}
Martin Heusel, Hubert Ramsauer, Thomas Unterthiner, Bernhard Nessler, and Sepp Hochreiter.
\newblock Gans trained by a two time-scale update rule converge to a local nash equilibrium.
\newblock \emph{Advances in neural information processing systems}, 30, 2017.

\bibitem[Ho et~al.(2020)Ho, Jain, and Abbeel]{ho2020denoising}
Jonathan Ho, Ajay Jain, and Pieter Abbeel.
\newblock Denoising diffusion probabilistic models.
\newblock \emph{Advances in neural information processing systems}, 33:\penalty0 6840--6851, 2020.

\bibitem[Huang et~al.(2021)Huang, Lim, and Courville]{huang2021variational}
Chin-Wei Huang, Jae~Hyun Lim, and Aaron~C Courville.
\newblock A variational perspective on diffusion-based generative models and score matching.
\newblock \emph{Advances in Neural Information Processing Systems}, 34:\penalty0 22863--22876, 2021.

\bibitem[Hyv{\"a}rinen and Dayan(2005)]{hyvarinen2005estimation}
Aapo Hyv{\"a}rinen and Peter Dayan.
\newblock Estimation of non-normalized statistical models by score matching.
\newblock \emph{Journal of Machine Learning Research}, 6\penalty0 (4), 2005.

\bibitem[Jolicoeur-Martineau et~al.(2021)Jolicoeur-Martineau, Li, Pich{\'e}-Taillefer, Kachman, and Mitliagkas]{jolicoeur2021gotta}
Alexia Jolicoeur-Martineau, Ke~Li, R{\'e}mi Pich{\'e}-Taillefer, Tal Kachman, and Ioannis Mitliagkas.
\newblock Gotta go fast when generating data with score-based models.
\newblock \emph{arXiv preprint arXiv:2105.14080}, 2021.

\bibitem[Karras et~al.(2019)Karras, Laine, and Aila]{karras2019style}
Tero Karras, Samuli Laine, and Timo Aila.
\newblock A style-based generator architecture for generative adversarial networks.
\newblock In \emph{Proceedings of the IEEE/CVF conference on computer vision and pattern recognition}, pages 4401--4410, 2019.

\bibitem[Karras et~al.(2022)Karras, Aittala, Aila, and Laine]{karras2022elucidating}
Tero Karras, Miika Aittala, Timo Aila, and Samuli Laine.
\newblock Elucidating the design space of diffusion-based generative models.
\newblock \emph{arXiv preprint arXiv:2206.00364}, 2022.

\bibitem[Kingma et~al.(2021)Kingma, Salimans, Poole, and Ho]{vdm}
Diederik~P Kingma, Tim Salimans, Ben Poole, and Jonathan Ho.
\newblock Variational diffusion models.
\newblock \emph{arXiv preprint arXiv:2107.00630}, 2021.

\bibitem[Kong et~al.(2023)Kong, Brekelmans, and {Ver Steeg}]{kong2023informationtheoretic}
Xianghao Kong, Rob Brekelmans, and Greg {Ver Steeg}.
\newblock Information-theoretic diffusion.
\newblock In \emph{International Conference on Learning Representations}, 2023.
\newblock URL \url{https://arxiv.org/abs/2302.03792}.

\bibitem[Kong et~al.(2024)Kong, Liu, Li, Yogatama, and Ver~Steeg]{konginterpretable}
Xianghao Kong, Ollie Liu, Han Li, Dani Yogatama, and Greg Ver~Steeg.
\newblock Interpretable diffusion via information decomposition.
\newblock In \emph{The Twelfth International Conference on Learning Representations}, 2024.

\bibitem[Krizhevsky(2009)]{krizhevsky2009learning}
Alex Krizhevsky.
\newblock Learning multiple layers of features from tiny images.
\newblock pages 32--33, 2009.
\newblock URL \url{https://www.cs.toronto.edu/~kriz/learning-features-2009-TR.pdf}.

\bibitem[Lee et~al.(2023)Lee, Kim, and Park]{lee2023codi}
Chaejeong Lee, Jayoung Kim, and Noseong Park.
\newblock Codi: Co-evolving contrastive diffusion models for mixed-type tabular synthesis.
\newblock \emph{arXiv preprint arXiv:2304.12654}, 2023.

\bibitem[Liu et~al.(2022)Liu, Ren, Lin, and Zhao]{liu2022pseudo}
Luping Liu, Yi~Ren, Zhijie Lin, and Zhou Zhao.
\newblock Pseudo numerical methods for diffusion models on manifolds.
\newblock \emph{arXiv preprint arXiv:2202.09778}, 2022.

\bibitem[Lu et~al.(2022)Lu, Zhou, Bao, Chen, Li, and Zhu]{lu2022dpm}
Cheng Lu, Yuhao Zhou, Fan Bao, Jianfei Chen, Chongxuan Li, and Jun Zhu.
\newblock Dpm-solver++: Fast solver for guided sampling of diffusion probabilistic models.
\newblock \emph{arXiv preprint arXiv:2211.01095}, 2022.

\bibitem[Matejka and Fitzmaurice(2017)]{dino_data}
Justin Matejka and George Fitzmaurice.
\newblock Same stats, different graphs: generating datasets with varied appearance and identical statistics through simulated annealing.
\newblock In \emph{Proceedings of the 2017 CHI conference on human factors in computing systems}, pages 1290--1294, 2017.

\bibitem[McAllester(2023)]{mcallester2023mathematics}
David McAllester.
\newblock On the mathematics of diffusion models.
\newblock \emph{arXiv preprint arXiv:2301.11108}, 2023.

\bibitem[Meng et~al.(2023)Meng, Rombach, Gao, Kingma, Ermon, Ho, and Salimans]{meng2023distillation}
Chenlin Meng, Robin Rombach, Ruiqi Gao, Diederik Kingma, Stefano Ermon, Jonathan Ho, and Tim Salimans.
\newblock On distillation of guided diffusion models.
\newblock In \emph{Proceedings of the IEEE/CVF Conference on Computer Vision and Pattern Recognition}, pages 14297--14306, 2023.

\bibitem[Nichol and Dhariwal(2021)]{nichol2021improved}
Alexander~Quinn Nichol and Prafulla Dhariwal.
\newblock Improved denoising diffusion probabilistic models.
\newblock In \emph{International Conference on Machine Learning}, pages 8162--8171. PMLR, 2021.

\bibitem[Oord et~al.(2018)Oord, Li, and Vinyals]{cpc}
Aaron van~den Oord, Yazhe Li, and Oriol Vinyals.
\newblock Representation learning with contrastive predictive coding.
\newblock \emph{arXiv preprint arXiv:1807.03748}, 2018.

\bibitem[Ouyang et~al.(2022)Ouyang, Xie, and Cheng]{ouyang2022improving}
Yidong Ouyang, Liyan Xie, and Guang Cheng.
\newblock Improving adversarial robustness by contrastive guided diffusion process.
\newblock \emph{arXiv preprint arXiv:2210.09643}, 2022.

\bibitem[Radford et~al.(2021)Radford, Kim, Hallacy, Ramesh, Goh, Agarwal, Sastry, Askell, Mishkin, Clark, et~al.]{clip}
Alec Radford, Jong~Wook Kim, Chris Hallacy, Aditya Ramesh, Gabriel Goh, Sandhini Agarwal, Girish Sastry, Amanda Askell, Pamela Mishkin, Jack Clark, et~al.
\newblock Learning transferable visual models from natural language supervision.
\newblock In \emph{International conference on machine learning}, pages 8748--8763. PMLR, 2021.

\bibitem[Rhodes et~al.(2020)Rhodes, Xu, and Gutmann]{rhodes2020telescoping}
Benjamin Rhodes, Kai Xu, and Michael~U Gutmann.
\newblock Telescoping density-ratio estimation.
\newblock \emph{Advances in neural information processing systems}, 33:\penalty0 4905--4916, 2020.

\bibitem[Salimans and Ho(2021)]{salimans2021progressive}
Tim Salimans and Jonathan Ho.
\newblock Progressive distillation for fast sampling of diffusion models.
\newblock In \emph{International Conference on Learning Representations}, 2021.

\bibitem[Shih et~al.(2024)Shih, Belkhale, Ermon, Sadigh, and Anari]{shih2024parallel}
Andy Shih, Suneel Belkhale, Stefano Ermon, Dorsa Sadigh, and Nima Anari.
\newblock Parallel sampling of diffusion models.
\newblock \emph{Advances in Neural Information Processing Systems}, 36, 2024.

\bibitem[Sohl-Dickstein et~al.(2015)Sohl-Dickstein, Weiss, Maheswaranathan, and Ganguli]{jaschaneq}
Jascha Sohl-Dickstein, Eric~A Weiss, Niru Maheswaranathan, and Surya Ganguli.
\newblock Deep unsupervised learning using nonequilibrium thermodynamics.
\newblock \emph{arXiv preprint arXiv:1503.03585}, 2015.

\bibitem[Song et~al.(2020{\natexlab{a}})Song, Meng, and Ermon]{ddim}
Jiaming Song, Chenlin Meng, and Stefano Ermon.
\newblock Denoising diffusion implicit models.
\newblock \emph{arXiv preprint arXiv:2010.02502}, 2020{\natexlab{a}}.

\bibitem[Song and Ermon(2019)]{song2019generative}
Yang Song and Stefano Ermon.
\newblock Generative modeling by estimating gradients of the data distribution.
\newblock \emph{Advances in Neural Information Processing Systems}, 32, 2019.

\bibitem[Song and Ermon(2020)]{song2020improved}
Yang Song and Stefano Ermon.
\newblock Improved techniques for training score-based generative models.
\newblock \emph{Advances in neural information processing systems}, 33:\penalty0 12438--12448, 2020.

\bibitem[Song et~al.(2019)Song, Garg, Shi, and Ermon]{song2019sliced}
Yang Song, Sahaj Garg, Jiaxin Shi, and Stefano Ermon.
\newblock Sliced score matching: A scalable approach to density and score estimation.
\newblock \emph{arXiv preprint arXiv:1905.07088}, 2019.

\bibitem[Song et~al.(2020{\natexlab{b}})Song, Sohl-Dickstein, Kingma, Kumar, Ermon, and Poole]{diffusion_sde}
Yang Song, Jascha Sohl-Dickstein, Diederik~P Kingma, Abhishek Kumar, Stefano Ermon, and Ben Poole.
\newblock Score-based generative modeling through stochastic differential equations.
\newblock \emph{arXiv preprint arXiv:2011.13456}, 2020{\natexlab{b}}.

\bibitem[Song et~al.(2023)Song, Dhariwal, Chen, and Sutskever]{song2023consistency}
Yang Song, Prafulla Dhariwal, Mark Chen, and Ilya Sutskever.
\newblock Consistency models.
\newblock \emph{arXiv preprint arXiv:2303.01469}, 2023.

\bibitem[Sugiyama et~al.(2012)Sugiyama, Suzuki, and Kanamori]{Sugiyama_Suzuki_Kanamori_2012}
Masashi Sugiyama, Taiji Suzuki, and Takafumi Kanamori.
\newblock \emph{Bibliography}, page 309–326.
\newblock Cambridge University Press, 2012.

\bibitem[Vincent(2011)]{vincent2011connection}
Pascal Vincent.
\newblock A connection between score matching and denoising autoencoders.
\newblock \emph{Neural computation}, 23\penalty0 (7):\penalty0 1661--1674, 2011.

\bibitem[Watson et~al.()Watson, Chan, Ho, and Norouzi]{watsonlearning}
Daniel Watson, William Chan, Jonathan Ho, and Mohammad Norouzi.
\newblock Learning fast samplers for diffusion models by differentiating through sample quality.

\bibitem[Xiao et~al.(2021)Xiao, Kreis, and Vahdat]{xiao2021tackling}
Zhisheng Xiao, Karsten Kreis, and Arash Vahdat.
\newblock Tackling the generative learning trilemma with denoising diffusion gans.
\newblock In \emph{International Conference on Learning Representations}, 2021.

\bibitem[Yadin et~al.(2024)Yadin, Elata, and Michaeli]{yadin2024classification}
Shahar Yadin, Noam Elata, and Tomer Michaeli.
\newblock Classification diffusion models: Revitalizing density ratio estimation.
\newblock 2024.

\bibitem[Yang et~al.(2023)Yang, Hwang, and Ye]{yang2023zero}
Serin Yang, Hyunmin Hwang, and Jong~Chul Ye.
\newblock Zero-shot contrastive loss for text-guided diffusion image style transfer.
\newblock \emph{arXiv preprint arXiv:2303.08622}, 2023.

\bibitem[Zhang and Chen(2021)]{zhang2021diffusion}
Qinsheng Zhang and Yongxin Chen.
\newblock Diffusion normalizing flow.
\newblock \emph{Advances in Neural Information Processing Systems}, 34:\penalty0 16280--16291, 2021.

\bibitem[Zhu et~al.(2022)Zhu, Wu, Olszewski, Ren, Tulyakov, and Yan]{zhu2022discrete}
Ye~Zhu, Yu~Wu, Kyle Olszewski, Jian Ren, Sergey Tulyakov, and Yan Yan.
\newblock Discrete contrastive diffusion for cross-modal and conditional generation.
\newblock \emph{arXiv preprint arXiv:2206.07771}, 2022.

\end{thebibliography}
}

\newpage
\clearpage

\appendix
\section{Derivations and Proofs}

\subsection{Score Connection}\label{app:score}

We derive the following relation. 
\begin{align}
\nabla_x \log p_\logsnr(\vx) = - \frac{\epshat(\vx, \logsnr)}{\ce} 
\end{align}

\newcommand{\Za}{Z_\logsnr}
To keep track of intermediate random variables and their associated distributions, we will use the cumbersome but more precise information theory notation where capitals represent a random variable and lowercase represents values. 
Define the channel that mixes the signal, $X$, with Gaussian noise as $\Za \equiv \cx X + \ce \capeps$ with $\capeps \sim \mathcal N(0, \mathbb I)$ and data distribution $p(X)$, $\logsnr$ represents the log of the Signal-to-Noise Ratio (SNR), and $\sigma(\cdot)$ is the sigmoid function. In this notation, the probability that a mixture distribution takes a value, $\vx$, would be written $p(\Za = \vx)$. 

We start by re-writing the left-hand side in new notation, and expand the definition based on the noisy channel model, using $\delta(\cdot)$ for the Dirac delta. 
\begin{align*}
    \nabla_x \log p(\Za = \vx) &=  1/ p(\Za = \vx) \nabla_x  p(\Za=\vx) \\
    &= 1/ p(\Za=\vx) \nabla_x  \int d\bar \vx d\eps ~~ \delta(\vx - \cx \bar \vx - \ce \eps) p(X=\bar \vx) p(\capeps=\eps) \\
    &=  1/ p(\Za=\vx)  \int d\bar \vx d\eps ~~ (\nabla_x  \delta(\vx - \cx \bar \vx - \ce \eps)) p(X=\bar \vx) p(\capeps=\eps) \\
    &=  1/ p(\Za=\vx)  \int d\bar \vx d\eps ~~ ((- 1 / \ce) \nabla_{\eps}  \delta(\vx - \cx \bar \vx - \ce \eps)) p(X=\bar \vx) p(\capeps=\eps) \\
    &=  1/ p(\Za=\vx)  \frac{1}{\ce} \int d\bar \vx d\eps ~~    \delta(\vx - \cx \bar \vx - \ce \eps) p(X=\bar \vx) \nabla_{\eps} p(\capeps=\eps) \\
    &=  - 1/ p(\Za=\vx)  \frac{1}{\ce} \int d\bar \vx d\eps ~~    \delta(\vx - \cx \bar \vx - \ce \eps) p(X=\bar \vx) \eps p(\capeps=\eps) \\
    &=  - 1/ p(\Za=\vx)  \frac{1}{\ce} \int d\eps ~~  p(\Za=\vx, \capeps=\eps)   \eps  \\
    &=  - \frac{1}{\ce} \int d\eps ~~  p(\capeps=\eps | \Za=\vx)   \eps  \\
    &=  - \frac{\mathbb E_{\eps \sim p(\capeps | \Za=\vx)} [  \eps ]}{\ce}  \\
    &= - \frac{\epshat(\vx, \logsnr)}{\ce} 
\end{align*}
In the second line we expand, and in the third we just move the gradient inside the integral. In the fourth line we use the chain rule to relate the gradient over $\vx$ to the gradient over $\eps$ (introducing a sign flip). In the fifth line we use integration by parts to move the gradient (second sign flip). Taking the gradient of the Gaussian in the sixth line gives our third sign flip, and the factor of $\eps$. We can conclude by writing the expression in terms of a conditional distribution, and relating that to the optimal denoiser in
Eq.~\ref{eq:opt}.

\subsection{Mixture Distribution Density}\label{app:mix_density}

In this section, we derive the expression that shows that the density of a continuum of Gaussian mixture distributions can be written in terms of the optimal denoiser, $\epshat$, for the data distribution.
\begin{align}
-\log p_\zeta(\vx) &= c +  \half \int_{-\infty}^{\infty} d\bar \logsnr ~\mathbb E_{p(\eps)} [ \norm{\eps - \sqrt{\frac{\sigma(-\bar \logsnr)}{\sigma(-\beta)}} \epshat(\cxbar \vx + \cebar \eps, \beta)}] \nonumber \\ 
\beta &\equiv \sigma^{-1} \left( \sigma(\zeta) \sigma(\bar \logsnr) \right)
\end{align}

As in the previous section, we will adopt information theory notation. 
If we define the optimal denoiser for the input distribution, $p_\zeta(\vx)$, with a subscript as $\epshat_\zeta(\cdot, \cdot)$, we can write the density analogously to Eq.~\ref{eq:density_simple}. 
\begin{align}
-\log p_\zeta(\vx) &= c +  \half \int_{-\infty}^{\infty} d\bar \logsnr ~\mathbb E_{p(\eps)} [ \norm{\eps - \epshat_\zeta(\cxbar \vx + \cebar \eps, \bar \logsnr)}]  
\end{align}
Note that we now have to keep track of two log SNR values. One indicates how much noise is added to the new ``data'' distribution, the other is how much noise we add and then try to remove with our denoiser. 
The goal is to relate $\epshat_\zeta$ to $\epshat$. 
We can formally write down the optimal solution using the relation in Eq.~\ref{eq:opt}. 
$$
\epshat_\zeta(\vx, \bar \logsnr) = \mathbb E_{\eps \sim p(\capeps | Z=\vx)} [\eps]
$$
Now, however, the noise channel is defined differently. 
The channel mixes the signal, $\bar \vx \sim p_\zeta(\bar X)$,
with Gaussian noise, $\bar \eps \sim \mathcal N(0, \mathbb I)$, as $Z \equiv \cxbar \bar X + \cebar \bar \capeps$. 
And the noisy variable, $\bar X = \sqrt{\sigma(\zeta)} X + \sqrt{\sigma(-\zeta)} \capeps$, where we must be careful to distinguish the two independent sources of Gaussian noise. 

We start by expanding definitions. 
\begin{align*}
    \epshat_\zeta(\vx_\zeta, \bar \logsnr) 
    &= \frac{1}{p(Z=\vx_\zeta)} \int d\bar \eps ~~ p(\capeps=\bar \eps, Z=\vx_\zeta) ~\bar \eps \\
    &= \frac{1}{p(Z=\vx_\zeta)} \int d\bar \eps d\bar \vx ~~ \delta(\vx_\zeta - \cxbar \bar \vx - \cebar \bar \eps) ~~ p(\bar X = \bar\vx) ~~ p(\bar\capeps = \bar\eps)  ~\bar \eps \\
    &= \frac{1}{p(Z=\vx_\zeta)} \int d\bar \eps d\bar \vx d\vx d\eps ~~ \delta(\vx_\zeta - \cxbar \bar \vx - \cebar \bar \eps) ~~ \delta(\bar \vx - \sqrt{\sigma(\zeta)} \vx - \sqrt{\sigma(-\zeta)} \eps) \\
    &~~~~~~~~~~~~~~~~~~ \cdot p(X=\vx) ~ p(\capeps=\eps) ~ p(\bar\capeps = \bar\eps)  ~\bar \eps \\
    &= \frac{1}{p(Z=\vx_\zeta)} \int d\eps d\bar \eps d\vx~~ \delta(\vx_\zeta - \cxbar (\sqrt{\sigma(\zeta)} \vx + \sqrt{\sigma(-\zeta)} \eps) - \cebar \bar \eps) \\
    &~~~~~~~~~~~~~~~~~~ \cdot p(X=\vx) ~ p(\capeps=\eps) ~ p(\bar\capeps = \bar\eps)  ~\bar \eps 
\end{align*}
Now do a change of variables, a 2-d rotation with: 
$$\bar \eps' = a \bar \eps + b \eps, \eps' = -b \bar \eps  + a \eps,$$
$$a= \sqrt{\sigma(\bar \alpha) \sigma(-\zeta) / (1-\sigma(\zeta) \sigma(\bar \alpha))} , b=\sqrt{\sigma(-\bar \alpha)/ (1-\sigma(\zeta) \sigma(\bar \alpha))}.$$
This change of variables leads to the following. 
\begin{align*}
    \epshat_\logsnr(\vx_\zeta, \bar \logsnr) 
    &= \frac{1}{p(Z=\vx_\zeta)} \int d\eps' d\bar \eps' d\vx~~ \delta(\vx_\zeta - \sqrt{\sigma(\zeta) \sigma(\bar \alpha)} \vx - \sqrt{1-\sigma(\zeta) \sigma(\bar \alpha)} \bar \eps' ) \\
    &~~~~~~~~~~~~~~~~~~ \cdot p(X=\vx) ~ p(\capeps'=\eps') ~ p(\bar\capeps' = \bar\eps') ~ (b \bar \eps' + a \eps') \\
    &= b \frac{1}{p(Z=\vx_\zeta)} \int d\eps' d\bar \eps' d\vx~~ \delta(\vx_\zeta - \sqrt{\sigma(\zeta) \sigma(\bar \alpha)} \vx - \sqrt{1-\sigma(\zeta) \sigma(\bar \alpha)} \bar \eps' )   \\
    &~~~~~~~~~~~~~~~~~~ \cdot p(X=\vx) ~ p(\capeps'=\eps') ~ p(\bar\capeps' = \bar\eps') ~ \bar \eps' \\
    &= b \frac{1}{p(Z=\vx_\zeta)} \int d\eps' d\bar \eps' d\vx~~ \delta(\vz - \sqrt{\sigma(\beta)} \vx - \sqrt{1-\sigma(\beta)} \bar \eps' ) ~ p(X=\vx) ~ p(\capeps'=\eps') ~ p(\bar\capeps' = \bar\eps') ~ \bar \eps'  \\
    \epshat_\zeta(\vx_\zeta, \bar \logsnr)  &= b \epshat(\vx_\zeta, \beta), \qquad \beta \equiv \sigma^{-1}(\sigma(\bar \alpha) \sigma(\zeta)), b=\sqrt{\sigma(-\bar \alpha)/ (1-\sigma(\zeta) \sigma(\bar \alpha))}
\end{align*}
Note in the second line that the expectation of $\eps'$ is zero, and we move the constant for the other term, $b$, outside the integral. 
In the third line, we define $\beta$ which represents the log SNR of the two consecutive noisy channels with $\zeta, \bar \alpha$. 
Then we recognize the resulting integral as Eq.~\ref{eq:opt}, the optimal denoiser for recovering samples from from the original (non-noisy) data distribution in Gaussian noise. 


\subsection{Main Plot 1D Two-mode Gaussian's Analytical Solution} \label{sec:1d_gauss_exp}

In this section, we calculate the analytical solution to the 1D two-mode Gaussian in Fig.~\ref{fig:main_plot}. 

Fig.~\ref{fig:main_plot} is plotting the norm of difference between the ground-truth denoiser $\epshat_{gt}(\cdot,\cdot)$ and the estimated denoiser $\epshat(\cdot,\cdot)$:
\begin{align*}
    \text{denoiser\_err}(\vx,\logsnr) = \norm{\epshat(\vx,\logsnr) - \epshat_{gt}(\vx,\logsnr)}
\end{align*}

To get this error plot, we need to analytically calculate ground-truth denoiser $\epshat_{gt}(\vx,\logsnr)$. 
From score connection Eq.~\ref{eq:score}, for any intermediate noisy density $\log p_\logsnr(\vx)$, denoiser function $\epshat(\vx,\logsnr)$ can be derived from score function $\nabla_x \log p_\logsnr(\vx)$. 
Therefore, ultimately what we need to calculate here is the score function of any noisy distribution $p_\logsnr(\vx)$. 

The data we used is a mixture of two Gaussians, $\mathcal N(\mu=-5,\mathbb I)$ and $\mathcal N(\mu=5,\mathbb I)$, and the noise distribution is consists of data plus noise, then the noisy distribution should also be a mixture of Guassians. We just need to relate the parameters of the noisy mixture of Gaussians to the parameters of the mixture of Gaussians. 

Start with one mode of the Gaussian mixture for the data. We could represent it in terms of the standard normal random variable, $\eps$. 
\begin{align*}
    \vx_d = \mu + \sigma \eps
\end{align*}
Now call $\vx_\logsnr$ the random variable after applying a noisy channel with log-SNR $\logsnr$, here we present sigmoid function as $\sigma(\cdot)$ .
\begin{align*}
    \vx_\logsnr = \cx \vx_d + \ce \eps'
\end{align*}
Note that we use a different $\eps'$ here. Now expand this, and then re-arrange.
\begin{align*}
    \vx_\logsnr &= \cx(\mu + \sigma\eps) + \ce\eps' \\
    &= \cx\mu + \cx\sigma\eps + \ce\eps'
\end{align*}
We want to represent this in a more canonical way to see what the variance and mean of this Gaussian is. Note that for two standard normal random variables, $a \eps + b \eps'$, we can represent them as a single random variable with the same variance, $\sqrt{a^2+b^2} \eps''$ (reparameterization trick).
\begin{align*}
    \vx_\logsnr = \cx \mu + \sqrt{\sigma(\logsnr)\sigma + \sigma(-\logsnr)}\eps''
\end{align*}
Now we see that the noisy Gaussian (one component of a mixture) is just a modified version of the original. We have to change the mean (moving it towards zero when adding noise) and the variance. 

In our example, we set $\sigma=1$, so it simplifies further.
\begin{align*}
    \vx_\logsnr = \cx \mu + \eps''
\end{align*}
So the variance doesn't change, we just slowly shift the two mixtures together to the center. 

Therefore, for one mode $\vx \sim \mathcal N(\mu,I)$, the intermediate noisy log-density is.
\begin{align*}
    \log p_\logsnr(\vx) = -\half \log(2 \pi \sigma^2) - \frac{(\vx - \cx\mu)^2}{2\sigma^2}
\end{align*}
Take the gradient of the log-density via torch built-in function torch.autograd.grad(log-density, samples), we have the ground-truth score function $\epshat_{gt}(\vx,\logsnr)$.

\section{Implementation Details}

\subsection{Synthetic Experiment -- Maximum Mean Discrepancy Bandwidth Choice} \label{app:mmd}

The Maximum Mean Discrepancy (MMD) is a statistical test used to determine if two distributions are different. 
It works by comparing the mean embeddings of samples drawn from two distributions in high dimensional feature space. Specifically, if the distributions are the same, the means should be close; if they are different, the means should be far apart. 
The embeddings are typically constructed using a feature map associated with a kernel function, and here we select the Gaussian kernel: 
\begin{equation*}
    K(\vx,\bm y) = \exp (-\frac{\| \vx - \bm y \|^2}{2\sigma^2})
\end{equation*}
The bandwidth parameter $\sigma$ of the Gaussian kernel plays a critical role in the sensitivity and performance of MMD. The better bandwidth choice, the more effective MMD computation is. 
Intuitively, the bandwidth $\sigma$ controls the scale at which differences between distributions are detected. A small $\sigma$ makes the kernel sensitive to differences at small scales (fine details), while a large $\sigma$ highlights differences at larger scales.

The choice of bandwidth is often related to the variance of the data, and the bandwidth should be on the order of the variance of the data. 
Through a small experiment where we calculate MMD score between our data and the standard Gaussian under various bandwidths $\sigma$s, we pick the one that maximizes the MMD score. 

\begin{figure}[h]
    \centering
    \includegraphics[width=0.5\textwidth]{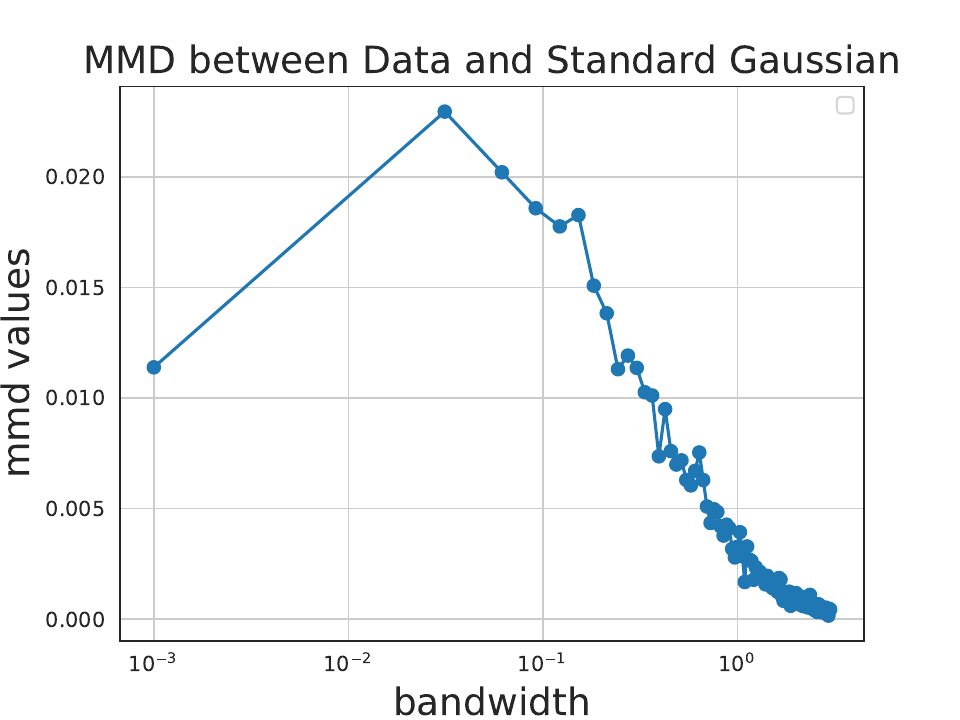}
    \caption{MMD between Dino data and the standard Gaussian. We plots the relationship between the MMD values and the bandwidth parameter used in the kernel function, and pick the bandwidth value with peak MMD score. }
    \label{fig:MMD_bandwidth}
\end{figure}

In the synthetic 2D Dino experiment, we plot the relationship between MMD scores and bandwidths (Fig.~\ref{fig:MMD_bandwidth}), setting $\sigma = 3e-02$. 

\subsection{Details on Model Training} \label{app:model_training}

\paragraph{Model Checkpoints} ~ 
We adopt two models as baselines to fine-tune with CDL: DDPM model provided by ~\citeauthor{ho2020denoising}~\citeyear{ho2020denoising} and EDM model provided by ~\citeauthor{karras2022elucidating}~\citeyear{karras2019style}. 

For DDPM model, the checkpoint \footnote{\url{https://github.com/pesser/pytorch_diffusion}} we used is a ema one pre-trained on unconditional CIFAR-10. The reason we are not using the most frequently used checkpoint (\url{https://huggingface.co/google/ddpm-cifar10-32}) is that, this is not EMA checkpoint and our calculation of the FID score of this model on $50,000$ generated image via sequential DDPM sampler gives $12.43$. This FID score is much higher than what reported on the original paper $3.17$.
Here we provide the FID scores of this non-EMA pre-trained model, results shown in Tab.~\ref{tab:non_ema_ddpm}. 
\begin{table}[h]
    \centering
    \begin{tabular}{c|cc}
        \toprule
             & Parallel DDPM Sampler & Sequential DDPM Sampler \\
        \midrule
        DDPM & $10.69 $        & $12.43$ \\
        CDL  & \bm{$7.83 $}    & \bm{$10.06$} \\
        \bottomrule
    \end{tabular}
    \caption{Evaluating FID score for both parallel and sequential DDPM samplers. FID scores are calculated using $5,0000$ samples.  }
    \label{tab:non_ema_ddpm}
\end{table}

For EDM model, in total eight checkpoints we used are “DDPM++ cont. (VP)” and “NCSN++ cont. (VE)” models pre-trained on three datasets (CIFAR-10, uncond-FFHQ, and uncond-AFHQv2 ~\cite{krizhevsky2009learning,karras2019style,choi2020stargan}) with two training settings (unconditional and conditional)
\footnote{\url{https://nvlabs-fi-cdn.nvidia.com/edm/pretrained/edm-afhqv2-64x64-uncond-ve.pkl}}
\footnote{\url{https://nvlabs-fi-cdn.nvidia.com/edm/pretrained/edm-afhqv2-64x64-uncond-vp.pkl}}
\footnote{\url{https://nvlabs-fi-cdn.nvidia.com/edm/pretrained/edm-cifar10-32x32-cond-ve.pkl}}
\footnote{\url{https://nvlabs-fi-cdn.nvidia.com/edm/pretrained/edm-cifar10-32x32-cond-vp.pkl}}
\footnote{\url{https://nvlabs-fi-cdn.nvidia.com/edm/pretrained/edm-cifar10-32x32-uncond-ve.pkl}}
\footnote{\url{https://nvlabs-fi-cdn.nvidia.com/edm/pretrained/edm-cifar10-32x32-uncond-vp.pkl}}
\footnote{\url{https://nvlabs-fi-cdn.nvidia.com/edm/pretrained/edm-ffhq-64x64-uncond-ve.pkl}}
\footnote{\url{https://nvlabs-fi-cdn.nvidia.com/edm/pretrained/edm-ffhq-64x64-uncond-vp.pkl}}. 

As for fine-tuning, we train all models with the same training setting in their original papers. 
For DDPM model, we train each model for 10 epochs and keep 'learning rate / batch size' ratio to be '$10^{-4}/64$', and this training is on two A6000 GPUs. 
For EDM model, the following table list the exact our fine-tuning configurations, which is still of the same 'learning rate/batch size' ratio. This training is using eight V100 GPUs. 

\begin{table}
    \centering
    \label{tab:configuration}
    \renewcommand{\arraystretch}{1.5} 
    \begin{tabularx}{\textwidth}{@{}l|X@{}} 
        \toprule
        \textbf{Dataset} & \textbf{Fine-tuning Configurations} \\
        \midrule
        uncond/cond CIFAR-10 & \texttt{--duration=0.5 --batch=128 --lr=2e-4} \\
        \midrule
        uncond AFHQ-64       & \texttt{--duration=0.5 --batch=32 --lr=5e-5 --cres=1,2,2,2 --dropout=0.25 --augment=0.15} \\
        \midrule
        uncond FFHQ-64       & \texttt{--batch=32 --lr=5e-5 --cres=1,2,2,2 --dropout=0.05 --augment=0.15} \\
        \bottomrule
    \end{tabularx}
    \caption{Fine-tuning configurations for different datasets}
\end{table}

\paragraph{More Experimental results with Parallel DDPM Sampler} ~
The previous parallel diffusion sampling paper~\cite{shih2024parallel} calculates FID scores by using $5,000$ generated images and another $5,000$ randomly selected real images, and to follow the same experimental setting for comparison, we further provide the FID results in Table~\ref{tab:parallel_sampling_5k}. 

\begin{table}[h]
    \centering
    \begin{adjustbox}{center}
    \small
    \begin{tabular}{@{}l|cccc@{}}  
        \toprule
        \multirow{2}{*}{Models} & \multicolumn{2}{c}{CIFAR-10 at 32x32} & AFHQv2 64x64 & FFHQ 64x64 \\
        \cmidrule(lr){2-5}
         & unconditional & conditional & unconditional & unconditional \\
        \midrule
        DDPM & $9.43$ & NA & NA & NA \\
        CDL-DDPM & \bm{$9.06$} & NA & NA & NA \\
        \midrule
        VP & $7.93 \pm 0.07$ & $7.67 \pm 0.07$ & $4.58 \pm 0.07$ & $6.26 \pm 0.07$ \\
        CDL-VP & \bm{$7.47 \pm 0.07$} & \bm{$7.27 \pm 0.07$} & \bm{$4.51 \pm 0.04$} & \bm{$5.89 \pm 0.07$} \\
        \midrule
        VE & $7.81 \pm 0.07$ & $7.59 \pm 0.07$ & $4.65 \pm 0.10$ & $6.33 \pm 0.07$ \\
        CDL-VE & \bm{$7.35 \pm 0.07$} & \bm{$7.19\pm0.07$} & \bm{$4.54 \pm 0.07$} & \bm{$5.94 \pm 0.07$} \\
        \bottomrule
    \end{tabular}
    \end{adjustbox}
    \caption{Evaluating FID score (lower is better) of parallel DDPM sampler on real-world datasets using $5,000$ samples. 
    “NA” stands for "Not Applicable". For reported FID scores, we run three sets of random seeds and reported the average with uncertainty.}
    \label{tab:parallel_sampling_5k}
    \vspace{-13pt}
\end{table}

\paragraph{More Details on EDM Fine-tuning} ~ 
As the design choices of EDM model is very comprehensive and complicate, here we list the training noise distribution, loss weighting, network and preconditioning choices we make during CDL fine-tuning in Tab.~\ref{tab:edm_finetune_choices}.

\begin{table}
    \centering
    \begin{tabular}{c|c}
        \toprule
        Network and preconditioning & \\
        \midrule
        Architecture of denoising function      & (any)\\
        Skip scaling $c_{skip}(\sigma)$         & $\sigma_{data}^2 / (\sigma^2 + \sigma_{data}^2)$\\
        Output scaling $c_{out}(\sigma)$        & $\sigma \cdot \sigma_{data} / \sqrt{\sigma_{data}^2 + \sigma^2}$\\
        Input scaling $c_{in}(\sigma)$          & $1 / \sqrt{\sigma^2 + \sigma_{data}^2}$\\
        Noise Cond. $c_{noise}(\sigma)$         & $\frac{1}{4} \ln(\sigma)$\\
        \midrule
        Training & \\
        \midrule
        Noise distribtion & $\ln(\sigma) \sim \mathcal{N}(P_{mean}, P_{std}^2)$\\
        Loss weighting $\lambda(\sigma)$ & $(\sigma^2 + \sigma_{data}^2) / (\sigma \cdot \sigma_{data})^2$\\
        \bottomrule
    \end{tabular}
    \caption{CDL finetune on EDM experiment -- fine-tuning design choices. }
    \label{tab:edm_finetune_choices}
    \vspace{-10pt}
\end{table}

\subsection{Relationship Among Log-SNR, Timesteps, and Noise variance Sigma}\label{app:sacling_relations}

To use the pre-trained models in the literature with our CDL loss, we need to translate "$t$", a parameter representing time in a Markov chain that progressively adds noise to data in ~\citeauthor{ho2020denoising}~\citeyear{ho2020denoising} and \citeauthor{diffusion_sde}~\citeyear{diffusion_sde}, or "$\sigma$", the variance scale of the Gaussian noise in ~\citeauthor{karras2022elucidating}~\citeyear{karras2022elucidating}, to a $\log$-SNR "$\logsnr$". 

\paragraph{Translation between Timesteps and Log-SNR} ~
For time-step $t$ in DDPM and stochastic diffusion notation, we recommend readers check \citeauthor{kong2023informationtheoretic}~\citeyear{kong2023informationtheoretic} Appendix B.2 about the mapping between $\logsnr$ and $t$. 

\paragraph{Translation between Noise Variance Sigma and Log-SNR} ~ 
For variance scale of the Gaussian noise $\sigma$ in EDM, referring to Eq.(7) and (8) in ~\citeauthor{karras2022elucidating}~\citeyear{karras2022elucidating}, it's easy to translate the pre-conditioning:
\begin{align}
\vx_{\logsnr} &\equiv c_{in}(\sigma) \cdot (\vx + \sigma \eps) \label{eq:edm_mixing}\\
\vx_\logsnr &\equiv \cx \vx + \ce \eps \equiv \cx ~ (\vx + \sqrt{\frac{\sigma(-\logsnr)}{\sigma(\logsnr)}} \eps) \label{eq:cdl_mixing}
\end{align}
From Eq.~\ref{eq:cdl_mixing} and Eq.~\ref{eq:edm_mixing}, we see that $\sigma \equiv \sqrt{\frac{\sigma(-\logsnr)}{\sigma(\logsnr)}}$, therefore, the relationship between $\logsnr$ and $\sigma$ should be:
$$ \sigma \equiv \exp(-\logsnr / 2), ~ \logsnr \equiv  -2\ln(\sigma) $$

\subsection{Contrastive Loss Implementation} \label{app:cdl_implementation}

To implement contrastive loss, we follow the definition in Sec.~\ref{sec:llr}. First, we generate a random binary label $y$. Next, conditioned on y, we sample from either data distribution $p(\vx)$ or the noisy data distribution $p_\zeta(\vx)$. 
We calculate the point-wise log-likelihood, then the contrastive loss in Eq.~\ref{eq:pointwise_cdl}. 

\begin{algorithm}[h]
    \caption{Contrastive Diffusion Loss -- Training}
    \label{algo:cdl_train}
    \begin{algorithmic}[1]
        \Repeat
            \State $x_0 \sim p(x_0)$
            \State $\zeta \sim \text{Uniform}(6, \ldots, 15)$
            \State \# uniformly sample from $p(x)$ or $p_\zeta(x)$
            \If{$\text{rand-prob} < 0.5$}
                \State $x = x_0$
                \State $y = 1$
            \Else
                \State $x = \text{generate\_mixture}(x_0, \zeta)$
                \State $y = -1$
            \EndIf
            \State \# calculate negative log-likelihood of $p(x), p_\zeta(x)$
            \State log\_px = -nll($x$)
            \State log\_px\_zeta = -nll($x, \zeta$)
            \State cdl\_loss = softmax($y\cdot$ (log\_px\_zeta - log\_px))
        \Until{converged}
    \end{algorithmic}
\end{algorithm}

\subsection{Training cost}~ \label{app:cost}
As we mentioned, CDL training is more expensive to compute than the standard diffusion loss, and here we analysis and give the reason. 

According to \citeauthor{kong2023informationtheoretic}~\citeyear{kong2023informationtheoretic}, we can write the pointwise standard diffusion loss function as Eq.~\ref{eq:pointwise_nllx}, and therefore the standard diffusion loss is as Eq.~\ref{eq:nll}. 
\begin{equation}\label{eq:pointwise_nllx}
    nll(\vx) = -\log p(\vx) = c + \half \int_{-\infty}^{\infty}  \mathbb E_{p(\eps)} [ \norm{\eps - \epshat(\vx_\logsnr, \logsnr)}]~ d\logsnr. 
\end{equation}
\begin{equation}\label{eq:nll}
    nll = \mathbb E_{p(\vx)}[ -\log p(\vx) ] = c + \half \int_{-\infty}^{\infty}  \mathbb E_{p(\eps)~p(\vx)} [ \norm{\eps - \epshat(\vx_\logsnr, \logsnr)}]~ d\logsnr. 
\end{equation}

To train the standard diffusion loss, we simply need to optimize Eq.~\ref{eq:nll} by all the training data. 
However, to train the contrastive diffusion loss, we need to estimate $nll(x)$ and $nll(x+\zeta)$ term in Algo.~\ref{algo:cdl_train}, and there we estimate Eq.~\ref{eq:pointwise_nllx} by duplicating a single data point $\vx$ for $N=100$ times and calculate Eq.~\ref{eq:nll}.  
This pointwise NLL estimation $nll(x)$ demands $N=100$ times more computational resources compared to the standard diffusion loss.

In principle, the contrastive loss Algo.~\ref{algo:cdl_train} should be executed for the entire training dataset. However, due to the high computational cost, we optimize only one data point per batch instead of utilizing all the training data. 

\section{Samples Visualization}\label{app:samplers}

We provide visualization of the images generated from pre-trained models fine-tuned via CDL loss. 

\begin{figure}[!ht]
    \centering
    \includegraphics[width=0.7\textwidth]{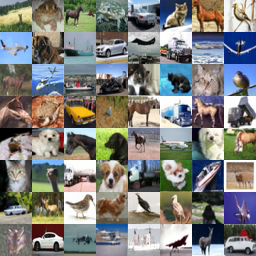}
    \caption{The CDL-loss fine-tuned EDM checkpoint generated examples from Conditional CIFAR-10, via parallel DDPM sampler.}
    \label{fig:para_cond_cifar10}
\end{figure}

\clearpage
\begin{figure}[!ht]
    \centering
    \includegraphics[width=0.7\textwidth]{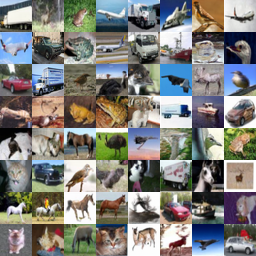}
    \caption{The CDL-loss fine-tuned EDM checkpoint generated examples from Unconditional CIFAR-10, via parallel DDPM sampler.}
    \label{fig:para_uncond_cifar10}
\end{figure}

\begin{figure}[!ht]
    \centering
    \includegraphics[width=0.7\textwidth]{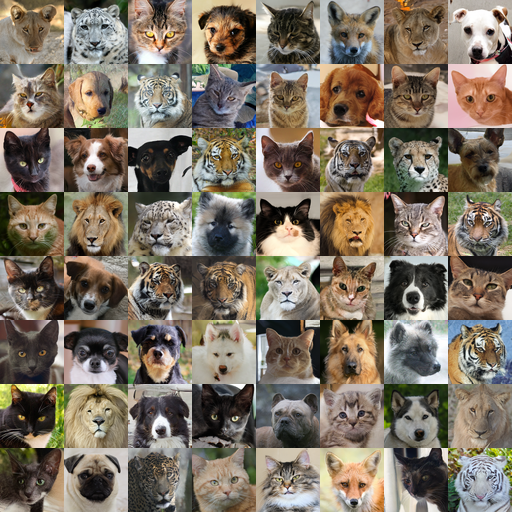}
    \caption{The CDL-loss fine-tuned EDM checkpoint generated examples from Unconditional AFHQ, via parallel DDPM sampler.}
    \label{fig:para_afhq}
\end{figure}

\begin{figure}[!ht]
    \centering
    \includegraphics[width=0.7\textwidth]{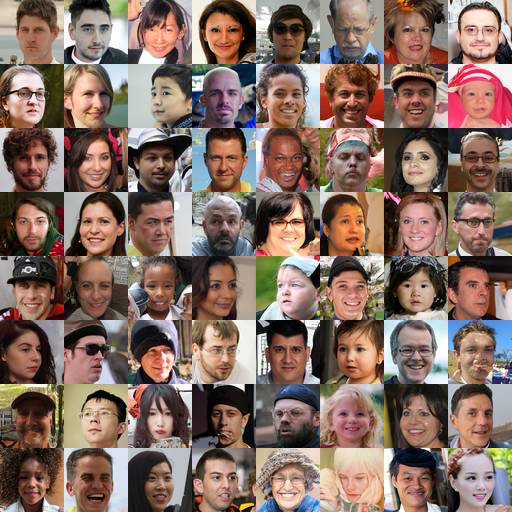}
    \caption{The CDL-loss fine-tuned EDM checkpoint generated examples from Unconditional FFHQ, via parallel DDPM sampler.}
    \label{fig:para_ffhq}
\end{figure}

\begin{figure}[!ht]
    \centering
    \includegraphics[width=0.7\textwidth]{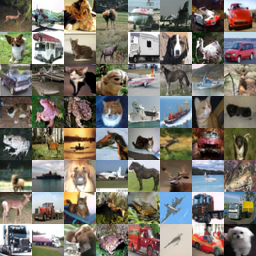}
    \caption{The CDL-loss fine-tuned EDM checkpoint generated examples from Conditional CIFAR-10, via sequential EDM sampler.}
    \label{fig: cond_cifar10}
\end{figure}

\clearpage

\begin{figure}[!ht]
    \centering
    \includegraphics[width=0.7\textwidth]{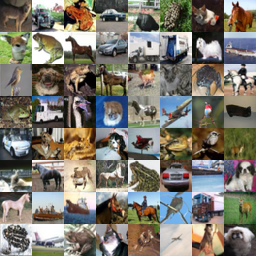}
    \caption{The CDL-loss fine-tuned EDM checkpoint generated examples from Unconditional CIFAR-10, via sequential EDM sampler.}
    \label{fig: uncond_cifar10}
\end{figure}

\begin{figure}[!ht]
    \centering
    \includegraphics[width=0.7\textwidth]{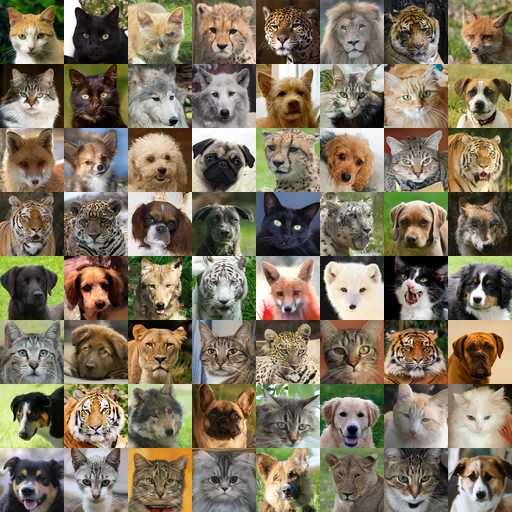}
    \caption{The CDL-loss fine-tuned EDM checkpoint generated examples from Unconditional AFHQ, via sequential EDM sampler.}
    \label{fig: afhq}
\end{figure}

\clearpage
\begin{figure}[!ht]
    \centering
    \includegraphics[width=0.7\textwidth]{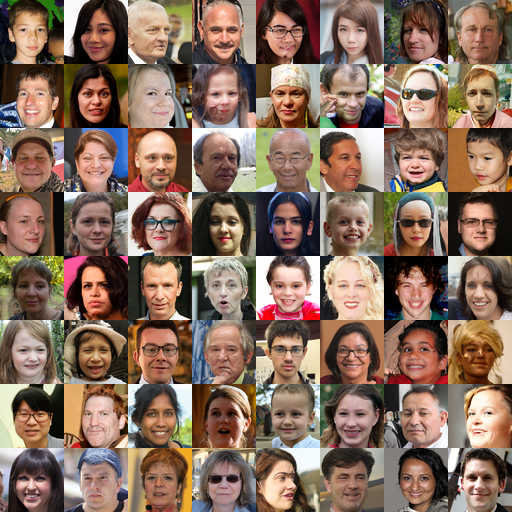}
    \caption{The CDL-loss fine-tuned EDM checkpoint generated examples from Unconditional FFHQ, via sequential EDM sampler.}
    \label{fig: ffhq}
\end{figure}

\clearpage 

\end{document}